\documentclass[lettersize,journal]{IEEEtran}
\usepackage{amsmath,amsfonts}
\usepackage{algorithmic}
\usepackage{algorithm}
\usepackage{array,threeparttable}
\usepackage[justification=centering]{caption}

\usepackage[caption=false,font=normalsize,labelfont=sf,textfont=sf]{subfig}
\usepackage{textcomp}
\usepackage{stfloats}
\usepackage{url}
\usepackage{verbatim}
\usepackage{graphicx}
\usepackage{cite}
\usepackage{amssymb}
\usepackage{amsmath}
\usepackage{amsthm}
\usepackage{mathtools}
\newtheorem{theorem}{Theorem}

\begin{document}

\title{Multi-Agent Deep Reinforcement Learning for Dynamic Avatar Migration in AIoT-enabled Vehicular Metaverses with Trajectory Prediction


}

\author{Junlong Chen, Jiawen Kang, Minrui Xu, Zehui Xiong, Dusit Niyato, \IEEEmembership{Fellow,~IEEE}, \\Chuan Chen, Abbas Jamalipour, \IEEEmembership{Fellow,~IEEE}, and Shengli Xie, \IEEEmembership{Fellow,~IEEE}
\thanks{Junlong Chen, Jiawen Kang, and Shengli Xie are with the school of Automation, Guangdong University of Technology
, Guangzhou, China (e-mail: 3121001036@mail2.gdut.edu.cn; kjwx886@163.com; Shlxie@gdut.edu.cn).}
\thanks{Minrui Xu and Dusit Niyato are with the school of Nanyang Technological University, Singapore, Singapore (e-mail: minrui001@e.ntu.edu.sg; DNIYATO@ntu.edu.sg).}
\thanks{Zehui Xiong is with the school of Singapore University of Technology and Design, Singapore, Singapore (e-mail: zehui\_xiong@sutd.edu.sg).}
\thanks{Chuan Chen is with the school of Sun Yat-sen University, Guangzhou, China (e-mail: chenchuan@mail.sysu.edu.cn).}
\thanks{Abbas Jamalipour is with the school of University of Sydney, Sydney
, Australia (e-mail: a.jamalipour@ieee.org).}}

\markboth{Journal of \LaTeX\ Class Files,~Vol.~14, No.~8, August~2021}{Shell \MakeLowercase{Chen \textit{et al.}}: A Sample Article Using IEEEtran.cls for IEEE Journals}

\maketitle

\begin{abstract}

Avatars, as promising digital assistants in Vehicular Metaverses, can enable drivers and passengers to immerse in 3D virtual spaces, serving as a practical emerging example of Artificial Intelligence of Things (AIoT) in intelligent vehicular environments.  The immersive experience is achieved through seamless human-avatar interaction, e.g., augmented reality navigation, which requires intensive resources that are inefficient and impractical to process on intelligent vehicles locally. Fortunately, offloading avatar tasks to RoadSide Units (RSUs) or cloud servers for remote execution can effectively reduce resource consumption. However, the high mobility of vehicles, the dynamic workload of RSUs, and the heterogeneity of RSUs pose novel challenges to making avatar migration decisions. To address these challenges, in this paper, we propose a dynamic migration framework for avatar tasks based on real-time trajectory prediction and Multi-Agent Deep Reinforcement Learning (MADRL). Specifically, we propose a model to predict the future trajectories of intelligent vehicles based on their historical data, indicating the future workloads of RSUs.Based on the expected workloads of RSUs, we formulate the avatar task migration problem as a long-term mixed integer programming problem. To tackle this problem efficiently, the problem is transformed into a Partially Observable Markov Decision Process (POMDP) and solved by multiple DRL agents with hybrid continuous and discrete actions in decentralized. Numerical results demonstrate that our proposed algorithm can effectively reduce the latency of executing avatar tasks by around 25\% without prediction and 30\% with prediction and enhance user immersive experiences in the AIoT-enabled Vehicular Metaverse (AeVeM).

\end{abstract}


\section{Introduction}
\subsection{Background and Motivations}

\IEEEPARstart{M}{etaverses}, as the immersive 3D Internet, is experiencing rapid development by providing immersive experiences to users through advanced technologies such as Augmented Reality (AR), Virtual Reality (VR), Digital Twin (DT), and Artificial Intelligence of Things (AIoT)~\cite{han2022dynamic}. Vehicular Metaverses \cite{huang2022joint}, as a new paradigm of intelligent transportation systems, aims to provide immersive virtual-physical interaction for drivers and passengers in a 3D virtual world through AIoT-enabled intelligent vehicles. Avatars, as crucial components of AeVeM, can serve as virtual assistants that can provide immersive, personalized, and customized interactive experiences for drivers and passengers.

Emerging vehicular avatar services, e.g., AR navigation, tourist guides, and 3D entertainment games~\cite{yao2021development}, require intensive computing and storage resources to perform high-quality, low-latency avatar tasks that provide drivers and passengers immersive experiences. AIoT, as a key enabler, plays a critical role in integrating intelligent vehicles with various IoT devices, cloud computing resources, and advanced AI algorithms, to enhance the overall system efficiency~\cite{liu2020toward}. Performing these tasks locally in intelligent vehicles with limited resources is usually inefficient and impractical~\cite{jiang2021reliable,li2022internet}. As an alternative solution, intelligent vehicles can upload avatar tasks to RSUs or cloud servers for real-time execution with low transmission latency \cite{zhu2022aerial}. Nevertheless, due to the high mobility of intelligent vehicles, the latency of offloading between intelligent vehicles and RSUs is changing dramatically. Specifically, when the distance between intelligent vehicles and RSUs is far, the communication latency becomes unacceptable, thus the continuity of the avatar service cannot be guaranteed \cite{chang20226g}. To maintain seamless high-quality avatar services, avatars need to be migrated online according to the movement of intelligent vehicles, leveraging the power of AIoT to optimize resource allocation and enhance user experiences~\cite{wu2023virtual}.

Empowered by the prediction of future routes of intelligent vehicles, avatar pre-migration refers to the migration of avatar tasks to the next RSU for pre-processing before the intelligent vehicles enter the coverage of current RSUs \cite{sun2015green}. The pre-migration can reduce the break-in-present of avatar services. However, due to the nature of the high mobility of intelligent vehicles, it is necessary to ensure the pre-migration of avatar tasks to RSUs with limited resources~\cite{wu2023social} and unpredictable workloads to reduce communication latency and waiting latency simultaneously without degrading the quality of immersion.
For vehicular avatar task migration, the challenges primarily revolve as follows.
\begin{itemize}

\item \textbf{Resource-intensive avatar tasks:} As the interactive avatar tasks require intensive resources for real-time reasoning and high-fidelity rendering~\cite{zhang2019analysis}, the availability of RSU computing resources in different locations of different regions might affect the completion of avatar tasks. Therefore, designing the optimal migration strategy for avatar tasks is crucial for minimizing latency.

\item \textbf{Dynamic RSU workload:} Dynamic variations in the topology of intelligent vehicle networks pose a challenge for predicting future RSU resource requirements. Inappropriate pre-migration decisions that do not take into account future RSU workloads can result in increased computational delays and reduced performance of avatar services, ultimately affecting user immersion. Therefore, pre-migration of avatar tasks to available RSUs based on future RSU workload can reduce service latency and provision online pre-migration for immersive experiences.

\item \textbf{Non-uniform RSU deployment:} The density of RSUs is uneven in different areas. In particular, the traffic of intelligent vehicles is high in urban areas, while it is low in remote areas, resulting in heterogeneous and non-uniform RSU deployment. Therefore, it is challenging to make adaptive decisions on which RSU to migrate and pre-migrate in advance.

\end{itemize}
\subsection{Solutions and Contributions}
To address these challenges, in this paper, we propose a dynamic avatar task migration framework to migrate avatars for seamless immersive experiences of users in AeVeM.
To address the challenges posed by the high mobility of intelligent vehicles and the non-uniform deployment of RSUs, we first propose a prediction model based on Long Short-Term Memory (LSTM)~\cite{hochreiter1997long} to output the future trajectories of intelligent vehicles. Specifically, historical trajectories of intelligent vehicles are input into the LSTM model for obtaining future routes of intelligent vehicles. Then, future routes of intelligent vehicles are further leveraged to predict the future locations of intelligent vehicles under the coverage of RSUs. Therefore, future workloads of RSUs can be estimated accurately. By accurately predicting the future workloads of RSUs, intelligent vehicles can proactively migrate to RSUs with lower workloads, leading to reduced system latency and improved performance.

In literature, existing work has shown that the avatar migration problem is a mixed-integer programming problem~\cite{sun2016primal} that is NP-hard and cannot be solved in a reasonable time using conventional algorithms. Moreover, the available resources of RSUs vary over time, and the migration decisions of different intelligent vehicles affect each other. To solve this problem, the problem is transformed into a Partially Observable Markov Decision Process. Fortunately, Multi-Agent Deep Reinforcement Learning (MADRL)~\cite{lowe2017multi} provides a feasible solution to the above problem. Accordingly, agents can learn strategies directly from interactions in high-dimensional environments and are often used in dynamic decision-making scenarios that maximize long-term returns~\cite{nguyen2020deep,zhang2023ai}. Therefore, in our proposed framework, we propose a MADRL-based algorithm to solve the task migration problem of avatars in AeVeM. By utilizing MADRL, our algorithm enables the avatars to learn and adapt their migration decisions based on the location of vehicles and available resources of RSUs that may vary over time. Furthermore, the migration decisions of different intelligent vehicles can influence each other, and MADRL allows for coordination and collaboration among multiple agents to achieve optimal migration strategies.

In contrast to MADRL with only one type of action space, MADRL with mixed action spaces can equip the agent with the appropriate type of decision space according to the properties of the control object, enabling more accurate decisions. There have been some works~\cite{liang2023hybrid, qiu2022hybrid, xu2022learning} that focused on investigating DRL with hybrid continuous and discrete action spaces to solve the resource allocation problem and achieved better results than DRL with a single type of action space. Inspired by these works, we propose a MAPPO-based scheme with hybrid action spaces, i.e., Hybrid-MAPPO, for the non-uniform RSU deployment and the resource-intensive avatar tasks. Hybrid-MAPPO allows each intelligent vehicle to execute the target pre-migrated RSU and the pre-migrated portion of the avatar task, which can further reduce the latency of avatar services.

The main contributions of this paper are summarized as follows.
\begin{itemize}
\item To provide users with an immersive experience of avatar services, we propose a novel framework for avatar task migration that takes into account the dynamic workload of RSUs, the heterogeneous deployment of RSUs, and the resource-intensive avatar task.

\item In contrast to conventional LSTM prediction models that do not take into account the RSU's coverage area where the intelligent vehicle is located, we propose a new coverage-aware LSTM prediction model in which the future location of intelligent vehicles in the RSU's coverage area and the RSU's future demands can be predicted based on the intelligent vehicles' historical trajectories. Numerous results demonstrate the model’s effectiveness in accurately estimating the future trajectories of intelligent vehicles, as indicated by a test Mean Squared Error (MSE) of $6.3*10^{-5}$.

\item We propose a new MAPPO-based scheme with hybrid action spaces that can adapt to the inconsistent deployment of RSUs and the resource-intensive avatar task. Numerous results show that our proposed scheme has better performance than the baseline schemes and can effectively reduce system latency by around 25\% without prediction and 30\% with prediction.

\end{itemize}

The remainder of this paper is organized as follows. Section \ref{related} discusses the related work. Section \ref{system} presents the system model. Section \ref{problem} shows the problem formulation. Section \ref{lstm} describes the LSTM-based trajectory prediction for avatar task pre-migration decisions. Section \ref{alg} describes the proposed multi-agent deep reinforcement learning scheme for pre-migration decisions. Section \ref{exp} demonstrates numerical results and the conclusion is presented in Section \ref{conclude}.

\section{Related Work}\label{related}
\subsection{Vehicular Metaverses}
The concept of Metaverses, initially introduced in the science fiction novel \textit{Snow Crash} as a parallel virtual space to the physical world, has captured public imagination and gained renewed attention following the success of the popular film \textit{Ready Player One}~\cite{jiang2021reliable}. These immersive experiences heavily rely on real-time rendering technologies like extended reality and spatial-sounding rendering, which serve as primary interaction interfaces but pose significant computational demands.

The rapid development of Metaverses has led to a surge of immersive applications, for example, online games, online virtual meetings, virtual shopping, and virtual travel~\cite{kang2022blockchain}. The authors in \cite{lim2022realizing} examined the challenges in Metaverse applications and explored the convergence of edge intelligence and Metaverse. In addition, the authors in \cite{xu2022full} explored the application of mobile edge networks in Metaverses and discussed the challenges of implementing Metaverses on resource-constrained edge devices. In \cite{jiang2021reliable}, the authors introduced the concept of vehicular Metaverses and proposed a framework that combined vehicular networks with edge and cloud computing to enable seamless and immersive intelligent vehicle services, such as AR navigation and 3D entertainment games. However, these works did not address the challenges of heterogeneous deployment of RSUs to achieve continuity of services in AeVeM.

\subsection{Service Migration}
Traditional migration of vehicle applications and services running on virtual machines, where virtual machines are transferred from one physical hardware unit to another to improve the quality of service or balance server workload, has already been studied in detail \cite{Osanaiye2017vm_migration}. For example, the authors in \cite{yu2018pre} proposed a dynamic VM migration scheduling scheme that reduced system latency by minimizing unnecessary VM migrations. With the rapid evolution of service migration, several studies have introduced the concept of digital twin migration. The authors in \cite{Lu2021dt_migration} proposed a digital twin migration framework for edge networks that used Deep Reinforcement Learning to reduce the latency of digital twin migration. The authors in \cite{sun2017adaptive} proposed an algorithm to implement avatar migration in cloudlet networks to reduce the end-to-end latency between UEs and their computational resources. However, these works did not consider avatar migration in AeVeM and the impact of intelligent vehicle mobility and RSU workload on avatar migration.

\subsection{Trajectory Prediction}
Accurate prediction of vehicle trajectories can improve traffic control and driving safety. The authors in \cite{Hou2013motion_model} presented a motion model-based vehicle trajectory prediction model to ensure the driving safety of autonomous vehicles. Inspired by the success of recurrent neural networks (RNNs) in predicting sequential data, the authors in \cite{zyner2019RNN} applied RNNs to vehicle trajectory prediction to predict the movement patterns of vehicles in cities, achieving higher accuracy than conventional trajectory prediction methods. Moreover, in \cite{guo2022v2v}, the authors employed a more advanced Long-Short-Term Memory (LSTM) model to predict vehicle trajectories to improve service quality. Unlike previous research, our study focuses on improving the quality of avatar services in vehicular Metaverse contexts by incorporating intelligent prediction of vehicle trajectories. To achieve this, we introduce a region-specific LSTM-based approach to vehicle trajectory prediction that incorporates spatial information, an aspect that has not been thoroughly explored in existing LSTM-based methods. This comprehensive framework not only enables more accurate trajectory predictions but also contributes to an improved user experience in vehicle Metaverses.

\begin{figure}[t]  	
\centerline{\includegraphics[width=0.5\textwidth]{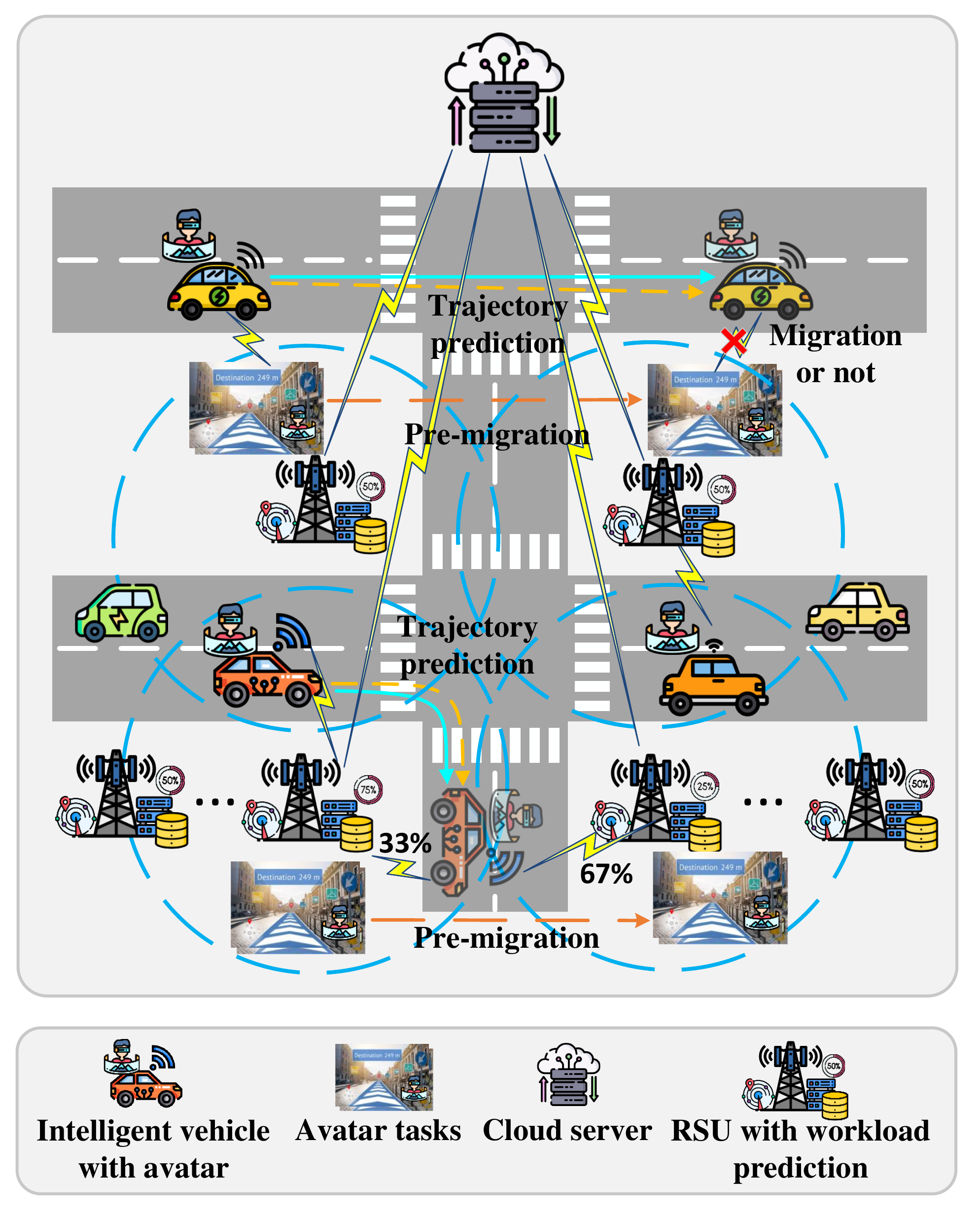}}  	
\caption{The migration of avatar tasks with hybrid action spaces and region-aware trajectory prediction in AeVeM.} \label{fig:system}  
\end{figure}

\begin{table}[t]

	\begin{threeparttable}
        \setlength{\abovecaptionskip}{0pt}
	\caption{Key symbols used in this paper}
	\begin{tabular}{c || p{6.5cm}}
		\hline
		\textbf{Notation} & \textbf{Description} \\
		\hline

        $B_{m,m_{p}}$ & The physical link bandwidth between source RSU $m$ and destination RSU $m_{p}$\\
        $B_{m,c}$ & The upload bandwidth of between RSU $m$ and cloud servers $c$\\

        $C_m$ &  The GPU computing resources of RSU $m$  \\
        $C_{cloud}$ & The GPU computing resources of cloud servers $c$\\
        $D_{m,v}^t$ & The distance between intelligent vehicle $v$ and RSU $m$\\
        $e_v$ &  The number of GPU cycles required per unit data of vehicle $v$  \\

        $K^{rd}_v(t)$ & The decisions of whether pre-migrating avatar task of intelligent vehicle $v$ in remote areas\\
        $K^{ud}_v(t)$ & The decisions of pre-migrating avatar task of intelligent vehicle $v$ to which RSU in urban areas\\
		$K^p_v(t)$ & The portion of pre-migrated avatar task of intelligent vehicle $v$\\

        $L_m(t)$ &  The current workload of RSU $m$  \\
		$L_m^{max}$ & The maximum workload of RSU $m$ \\
  	$\mathcal{M}$,$M$ & The set of RSUs and the number of RSUs\\

        $R^{up}_{m,v}(t)$ &  The uplink rate between intelligent vehicle $v$ and RSU $m$ \\
		$R^{d}_{m,v}(t)$ &  The downlink rate between intelligent vehicle $v$ and RSU $m$ \\
		$R^{c}_{v}(t)$ & The downlink rate between intelligent vehicle $v$ and cloud servers $c$\\

		$s_v^{task}(t)$ & The size of avatar task of intelligent vehicle $v$\\ 
		$s^{mig}_v(t)$ & The size of pre-migrated avatar task of intelligent vehicle $v$\\
		$s^{c}_{v}(t)$ & The size of avatar task of intelligent vehicle $v$ processed at the cloud servers $c$\\
        $t$ & Time slot \\
        $\mathcal{T}$, $T$ & The set of time slots and the max time slot \\
		$T^{up}_{m,v}(t)$ & The latency experienced by offloading an avatar task input of intelligent vehicle $v$ to RSU $m$\\
		$T^{d}_{v}(t)$ & The latency experienced by receiving avatar task results of intelligent vehicle $v$\\
		$T^{mig}_{v}(t)$ & The latency generated by pre-migrating avatar task of intelligent vehicle $v$ between RSUs \\
		$T_{m,v}^{dur}(t)$ & The duration time of intelligent vehicle $v$ in the coverage region of RSU $m$\\
		$T^{pro}_{m,v}(t)$ & The processing latency from receiving avatar task of intelligent vehicle $v$ to complete at RSU $m$\\
        $T^{pro}_{m_p,v}(t)$ & The processing latency from receiving avatar task of intelligent vehicle $v$ to complete at destination RSU $m_{p}$\\
        $T^{c}_{m,v}(t)$ & The processing latency from uploading task of intelligent vehicle $v$ to complete at cloud servers $c$\\
		$T^{sum}_v(t)$ &Total latency of avatar services of intelligent vehicle $v$ \\
		$\mathcal{V}$, $V$ & The set of intelligent vehicles and the number of intelligent vehicles\\

		\hline
	\end{tabular}
	\end{threeparttable}
	\label{tab:shape-functions}
\end{table}

\section{System Model} \label{system}

In this section, we present the avatar migration system of intelligent vehicles in AeVeMs. First, we present the network model, the migration model, and the computing model. Then, we introduce the total latency of avatar task migration in these models, respectively.
In AeVeM, the system consists of multiple RSUs and multiple intelligent vehicles, where the set of RSUs is denoted as $\mathcal{M}={\{1,\dots, m, \dots, M\}}$ and the set of intelligent vehicles is denoted as $\mathcal{V}={\{1,\dots, v,\dots, V\}}$. As shown in Fig.~\ref{fig:system}, the density of RSUs in different areas is non-uniform. For example, in urban areas, RSUs are densely deployed in high-traffic regions, while in remote areas, RSUs are sparsely deployed along highways. In AeVeM, RSU $m$ possesses GPU computing resources $C_m$, uplink bandwidth $B_m^{up}$, and downlink bandwidth $B_m^{d}$ for processing computation tasks, receiving avatar task inputs, and transmitting results of avatar task, respectively. The maximum workload capacity of RSU $m$ is denoted as $L_m^{max}$. In the system, we divide time into discrete time slots $\mathcal{T}=\{1,\dots, t,\dots, T\}$. Therefore, intelligent vehicles can transmit avatar task input (e.g., information required for AR navigation) to RSUs at time slot $t$ for remote execution. Alternatively, intelligent vehicles can proactively transfer a portion of avatar tasks to available RSUs to process in advance. To improve the rationality of pre-migration decisions, the system can also predict the future workload of RSUs based on the trajectories of intelligent vehicles, which in turn predicts the future locations of regions covered by RSUs.

\subsection{Network Model}
We first calculate the communication latency caused by avatar tasks via wireless transmission between intelligent vehicles and RSUs. During the wireless transmission, the position of RSU $m$ is considered to be fixed, which can be denoted as $P_m = (x_m,y_m)$. The position of intelligent vehicle $v$ varies with time, denoted as $P_v^t = (x_v^t,y_v^t)$ at time slot $t$. Therefore, the Euclidean distance between RSU $m$ and intelligent vehicle $v$ at time slot $t$ can be expressed as $D_{m,v}^t = \sqrt{|x_m-x_v^t|^2+|y_m-y_v^t|^2}$, where $|\cdot|$ represents the Euclidean distance operator \cite{Lee1993Euclid}.

In wireless communication, the latency generated by an intelligent vehicle $v$ offloading an avatar task input to RSU $m$ at time slot $t$ depends on the uplink rate. We consider the uplink rate~\cite{shannon2001mathematical}  
\begin{equation}
R^{up}_{m,v}(t)=B_m^{up}\log_2(1+\frac{p_v h_{m,v}^t}{\sigma^2_m}),\label{eq2}
\end{equation}
where $\sigma_m$ is the additive Gaussian white noise at RSU $m$, $p_v$ denotes the transmit power of intelligent vehicle $v$, and $h_{m,v}^t$ represents the wireless uplink channel and downlink channel between intelligent vehicle $v$ and RSU $m$ at time slot $t$. In this paper, we consider the Rayleigh fading channel, which can be calculated as $h_{m,v}^t = A(\frac{l}{4\pi f D_{m,v}^t})^2$, where $A$ denotes the channel gain coefficient, $l$ is the speed of light, $f$ denotes the carrier frequency, and $D_{m,v}^t$ is the Euclidean distance between intelligent vehicle $v$ and RSU $m$ at time slot $t$. Before users can enjoy immersive avatar services, intelligent vehicle $v$ needs to offload an avatar task input of size $s_v^{r}(t)$ to the RSU that is currently providing the service based on the user's demands.
Therefore, the transmission latency experienced by offloading an avatar task input of intelligent vehicle $v$ at time slot $t$ can be calculated as $T^{up}_{m,v}(t)=\frac{s_v^{r}(t)}{R^{up}_{m,v}(t)}$.

Unlike uploading an avatar task input to a specified RSU, the intelligent vehicle may receive the results of the completed avatar task from multiple RSUs or cloud servers. Specifically, when an avatar task input arrives at the RSU, the RSU generates an avatar task of size $s^{task}_v(t)$ according to the avatar task input. When the intelligent vehicle chooses to pre-migrate a portion of the avatar task to the specified RSU $m_{p}$, a portion of the avatar task with size $s^{mig}_v(t)$ can be migrated to that RSU. Furthermore, when avatar tasks cannot be completed within the deadline, the unfinished portion with size $s^{c}_{v}(t)$ is transferred to cloud servers with more computing resources for instant processing. After the cloud servers finish the task processing, the results of the processed avatar tasks are returned directly to the intelligent vehicle via the vehicular network. Similar to the uplink rate, the downlink rate at which intelligent vehicle $v$ receives data back from RSUs or cloud servers is defined as 
 \begin{equation}
R^{d}_{m,v}(t)=B_m^{d}\log_2(1+\frac{p_v h_{m,v}^t}{\sigma^2_m}).\label{eq3}
\end{equation}
Therefore, the transmission latency experienced by receiving the avatar task results of intelligent vehicle $v$ returned from RSUs or cloud servers at time slot $t$ is defined as
\begin{equation}
T^{d}_{v}(t)=\frac{s^{task}_v(t)-s^{mig}_v(t)}{R^{d}_{m,v}(t)} + \frac{s^{mig}_v(t)}{R^{d}_{m_{p},v}(t)} +\frac{s^{c}_{v}(t)}{R^{c}_{v}(t)},\label{eq4}
\end{equation}
where $s^{c}_{v}(t)$ is the size of the avatar task processed at cloud servers and $R^{c}_{v}(t)$ is the downlink transmission rate between intelligent vehicles and cloud servers.

\begin{figure}[t]  	
\centerline{\includegraphics[width=1\linewidth]{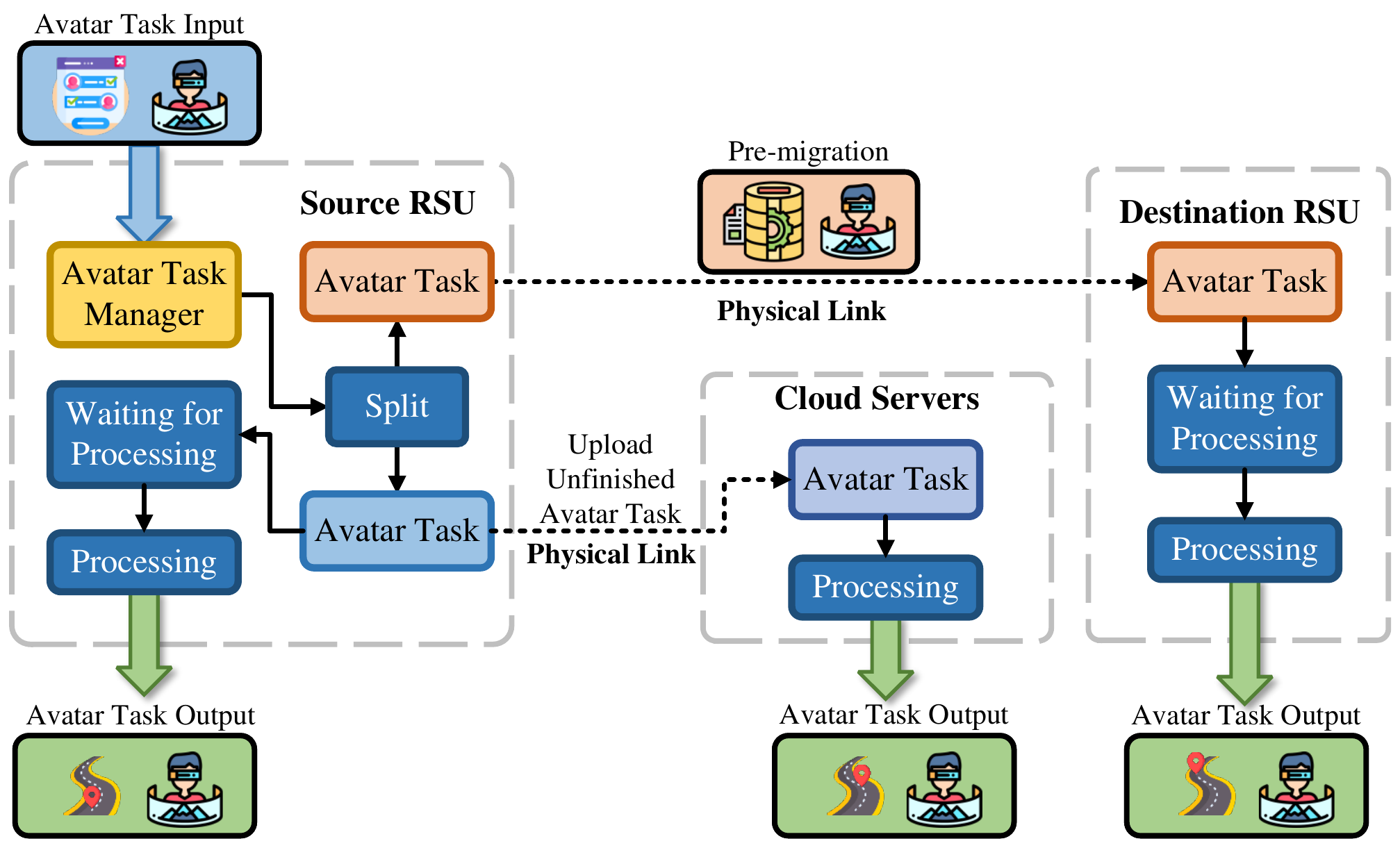}}  	
\caption{Process of avatar task migration.} \label{fig:pipeline}  
\end{figure} 

\subsection{Migration Model}
In the avatar task migration model, the avatar needs to be migrated to follow the vehicle when the location of the intelligent vehicle changes. When an intelligent vehicle leaves the region of the RSU that it currently serves, the avatar of that vehicle is migrated to the next RSU to enable seamless avatar services. Consequently, the RSUs to which the intelligent vehicle forwards avatar service inputs also change.  As shown in Fig.~\ref{fig:pipeline}, at time slot $t$, when an intelligent vehicle unloads an avatar task input, the avatar task pre-migration decision must be made simultaneously. Due to the dynamic workload of RSUs, non-uniform deployment of RSUs, and resource-intensive avatar tasks, we consider avatar task pre-migration decisions with hybrid action spaces. Specifically, the avatar task pre-migration decision $\mathcal{K}=\{K^{rd}_v(t)/K^{ud}_v(t),K^p_v(t)\}$ is divided into two components, i.e., the selection of the RSU to which to pre-migrate avatar tasks and the pre-migrated portion of the avatar tasks.

As shown in Fig.~\ref{fig:system}, the use of RSUs varies in different regions. In remote areas, RSUs are sparsely distributed and the workload capacity of RSUs is limited. Therefore, intelligent vehicles can often only decide whether to perform pre-migration or not. In the meanwhile, in urban areas, the number of RSUs is very high, so intelligent vehicles have more opportunities to pre-migrate avatar tasks to the RSU in that area. Therefore, we define the first part of the avatar task pre-migration decision as a binary decision, i.e., $K^{rd}_v(t)\in\{0,1\},\forall v\in \mathcal{V}$ (e.g., in remote areas) or as an integer decision, i.e., $K^{ud}_v(t)\in\{1,\dots,n\},\forall v\in \mathcal{V}$ (e.g., in urban areas), where $n$ is the number of RSUs that intelligent vehicle $v$ can currently select for the avatar pre-migration task.

The workloads of the different RSUs are different and the workloads of the RSUs are variable in time. Therefore, the unused resources of the RSU must be fully utilized to reduce the latency of the avatar services. It is not enough to just make decisions before migration but to optimize the dynamic portions of the avatar tasks before migration according to the dynamic nature of the RSU resources. Specifically, we define the second part of the migration decision for avatar tasks as $K^p_v(t)\in[0,1],\forall v\in \mathcal{V}$. Therefore, at time slot $t$ for intelligent vehicle $v$, the pre-migration avatar task size can be calculated as $s^{mig}_v(t)= K^p_v(t) s^{task}_v(t)$.

As shown in Fig.~\ref{fig:pipeline}, when the intelligent vehicle makes the decision to pre-migrate the avatar task, the avatar task is pre-migrated from the current RSU to the specified RSU via the physical link between RSUs. The bandwidth of the physical link between source RSU $m$ and destination RSU $m_{p}$ is defined as $B_{m,m_{p}}$. Therefore, the latency caused by pre-migrating avatar tasks between RSUs can be calculated as $T^{mig}_{v}(t)=\frac{s^{mig}_v(t)}{B_{m,m_{p}}}$. 

\subsection{Computation Model}
The migration model describes how avatar tasks are migrated between RSUs based on the location of the intelligent vehicle. In this section, we describe the avatar task computation model, which includes avatar task processing in both RSUs and cloud servers. As shown in Fig.~\ref{fig:pipeline}, the avatar tasks that have not been pre-migrated are processed after the intelligent vehicles make the avatar pre-migration decision, after the other avatar tasks have been processed in the current RSU. Therefore, at time slot $t$, the processing latency experienced by the avatar task of intelligent vehicle $v$ at the current RSU $m$ from waiting to be processed to the completion of the processing can be calculated as
\begin{equation}
T^{pro}_{m,v}(t)= \frac{L_m(t) + [1-K^p_v(t)] s^{task}_v(t) e_v}{C_m},\label{eq7}
\end{equation}
where $L_m(t)$ denotes the current workload of RSU $m$, $e_v$ represents the number of GPU cycles required per unit data of vehicle $v$ and $C_m$ signifies the GPU computing resources of RSU $m$ \cite{ren2020edge}.

Similarly, after the avatar task starts its pre-migration, the processing latency experienced by intelligent vehicle $v$ on RSU $m_{p}$ from the pre-migration of avatar tasks to the completion of processing at time slot $t$ can be calculated as
\begin{equation}
T^{pro}_{m_{p},v}(t)= T^{mig}_v(t) + \frac{L_{m_{p}}(t) + s^{mig}_v(t) e_v}{C_{m_{p}}}.\label{eq8}
\end{equation}

The time that intelligent vehicle $v$ is within the sensing range of RSU $m$ is defined as $T_{m,v}^{dur}(t)$. If the avatar task cannot be uploaded and processed within $T_{m,v}^{dur}(t)$, the remaining task of intelligent vehicle $v$ at RSU $m$ is uploaded to the cloud server for processing. Therefore, the size of the remaining task, i.e., the avatar task that will be computed at the cloud server, can be calculated as
\begin{equation}
\begin{aligned}
s^{c}_{v}(t)=
\begin{cases}
0, \quad \text{if} \quad T^{up}_{m,v}(t) + T^{pro}_{m,v}(t)\leq  T_{m,v}^{dur}(t), \\
\begin{aligned}
\frac{L_m(t) + [1-K^p_v(t)] s^{task}_v(t) e_v
- C_m T_{m,v}^{dur}(t)}{e_v},
\end{aligned}  \\ \text{~~~~~~~~~~~~~~~~~~~~~~~~~~~~~~~~~~~~~~~~~~~~~~otherwise.}
\end{cases}
\end{aligned}
\end{equation}

The remaining tasks can be uploaded to cloud servers and computed through the physical link between the RSU and cloud servers, which introduces additional cloud processing latency, is calculated as
\begin{equation}
T^{c}_{m,v}(t)= \frac{s^{c}_{v}(t)}{B_{m,c}} + \frac{s^{c}_{v}(t) e_v}{C_{cloud}},\label{eq10}
\end{equation}
where $B_{m,c}$ is the upload bandwidth of the physical link between RSU $m$ and cloud servers $c$, and $C_{cloud}$ is the GPU computing resources of the cloud servers.

\subsection{Total Latency of Avatar Tasks}
Overall, for AeVeM, when avatar services are provided, e.g., AR navigation, the intelligent vehicle offloads an avatar task input to the RSU via wireless communication. The uplink latency $T^{up}_{m,v}(t)$ is generated during the transmission of the input data. Once the avatar task input is uploaded, the RSU starts waiting for the avatar task to be processed in the current RSU, which causes a processing latency $T^{pro}_{m,v}(t)$ in the current RSU. At the same time, based on the intelligent vehicle's  pre-migration decision, the RSU migrates a certain size of the avatar task to a certain RSU, resulting in a migration latency $T^{mig}_v(t)$. After the pre-migration of the avatar task is complete, a processing latency $T^{pro}_{m_{p},v}(t)$ arises in the pre-migration RSU due to waiting for processing to complete.

Since the process of waiting for avatar task processing and pre-migration of avatar tasks are performed simultaneously, the latency generated by this process can be calculated as follows

\begin{equation}
T^{pro}_{v}(t)= \max \{T^{pro}_{m,v}(t),T^{pro}_{m_{p},v}(t) \}.\label{eq11}
\end{equation}

However, for the remaining tasks that cannot be processed before the intelligent vehicle leaves the region of the RSU, there is a cloud processing latency $T^{c}_{m,v}(t)$ due to processing in the cloud server. When the intelligent vehicle receives the result returned from the avatar tasks after the processing is complete, there is a downlink latency $T^{d}_{v}(t)$ of the RSUs and cloud servers.
Therefore, the total latency of the avatar service $T^{sum}_v(t)$ can be calculated as
\begin{equation}
T^{sum}_v(t)=T^{up}_{m,v}(t)+T^{pro}_{v}(t)+T^{c}_{m,v}(t)+T^{d}_{v}(t).\label{eq18}
\end{equation}
To minimize the average latency of avatar services, we formulate an optimization problem in the following section.
\section{Problem Formulation} \label{problem}
The objective of this system is to minimize the average latency of avatar services for all intelligent vehicles within a given finite time horizon $T$, under the constraint of maximum RSU utilization, by finding the optimal pre-migration decision policy defined as follows:
\begin{subequations}
\begin{align} 
\min_{\mathcal{K}} \quad &\sum_{t=1}^{T}\sum_{v=1}^{V} T^{sum}_v(t) \label{consta} \\
\text{s.t.} \quad &L_m(t) \leq L^{max}_m, \quad &&\forall m\in \mathcal{M}, \label{constb} \\
&L_{m}^{mig}(t) \leq L^{max}_{m_{p}}, \quad &&\forall m_{p}\in \mathcal{M}, \label{constc} \\
&K^{rd}_v(t)\in\{0,1\},\quad &&\forall v\in \mathcal{V}, \label{constd} \\
&K^{ud}_v(t)\in\{1,\dots,m,\dots,n\},\quad &&\forall v\in \mathcal{V}, \label{conste} \\
&K^p_v(t)\in[0,1],\quad &&\forall v\in \mathcal{V}, \label{constf} \\
&t\in {1, \ldots, T}. \label{constg}\\ \nonumber
\end{align}
\end{subequations}
Constraints (\ref{constb}) and (\ref{constc}) ensure that the workload of an RSU does not exceed its maximum capacity at any time. Constraints (\ref{constd}) and (\ref{conste}) guarantee that each avatar task of the vehicle can be assigned by one and only one RSU. Constraint (\ref{constf}) guarantees that the pre-migrating portion of the avatar task is less than the size of the avatar task. Constraint (\ref{constg}) states that the optimization problem takes place within a finite time. However, since the optimization problem in \eqref{eqproof1} is NP-hard, conventional algorithms have difficulty finding a feasible solution within an acceptable time. We present an NP-hard proof of this problem as follows.

\begin{theorem}
The avatar migration problem in~Eq. (\ref{consta}) is NP-hard.
\end{theorem}

\begin{proof}
To demonstrate that this problem is NP-hard, we first introduce the binary knapsack problem as~\cite{KexinLi2023nphard}:
\begin{equation}
\begin{aligned} \label{eqproof1}
&\max \quad \sum_{i=1}^{n} \omega_i  x_i \\
&\begin{array}{r@{\quad}r@{}l@{\quad}l}
s.t. & \sum_{i=1}^{n} v_i x_i \leq W ,x_i\in\{0,1\}.\\   
\end{array}
\end{aligned}
\end{equation}

In this problem, $n$ denotes the number of objects, $W$ denotes the capacity of the knapsack, while $\omega_i$ and $v_i$ represent the weight and volume of the $i$-th item correspondingly. The optimization variable in this problem is $x_i$, which is a binary decision variable indicating whether the $i$-th item is placed in the knapsack or not. The optimization objective, which is a linear function, is to maximize the total weight of the items in the knapsack. By mapping the variables in the binary knapsack problem to $v_i$,$x_i$, and $C$ in our proposed avatar services latency optimization problem \eqref{consta}, we obtain that the single slot optimization of avatar task migration is equivalent to the binary knapsack problem. Thus, the long-term latency minimization problem of avatar services is also NP-hard.
\end{proof}

\begin{figure}[t]
\centerline{\includegraphics[width=0.45\textwidth]{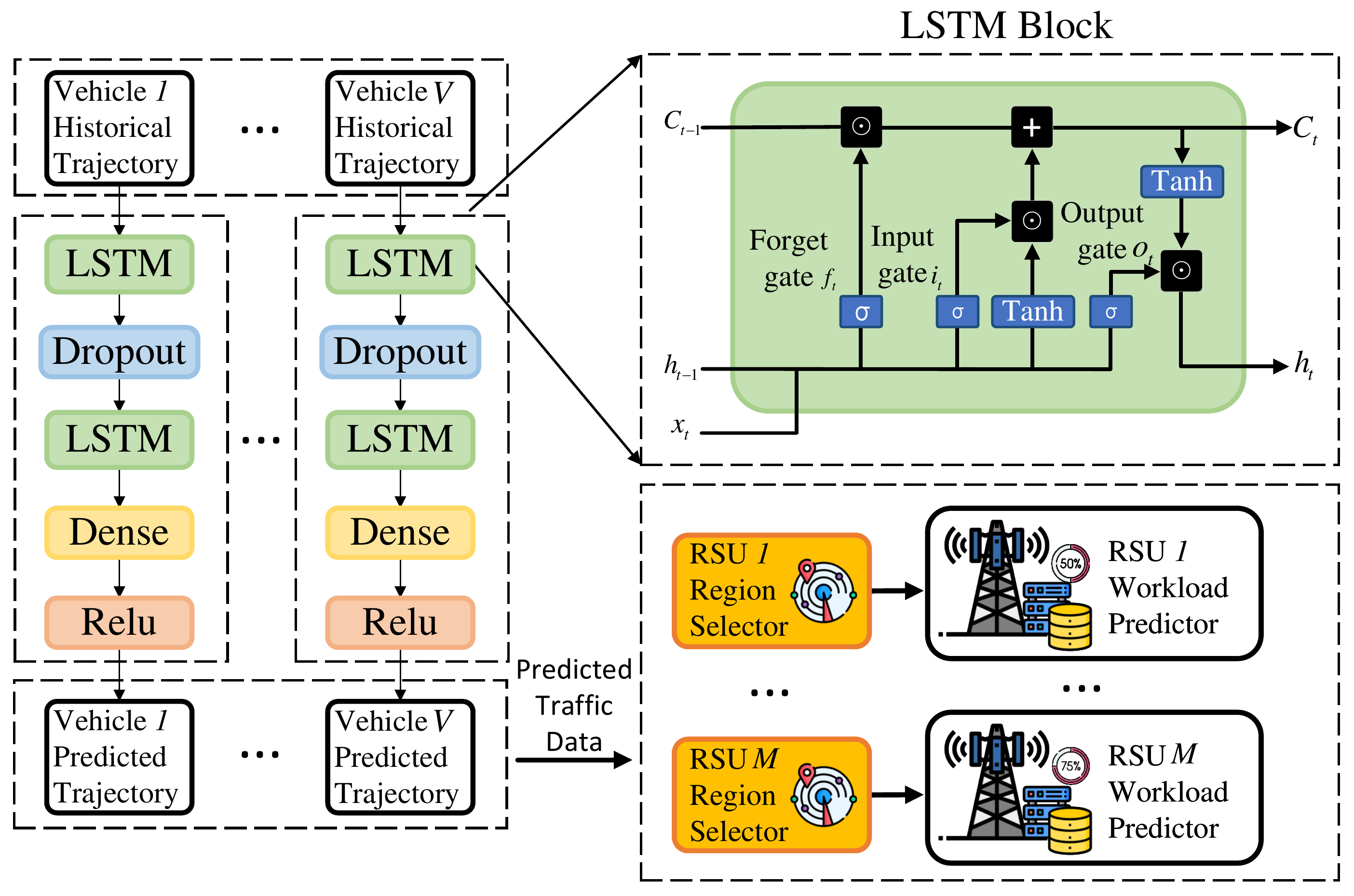}}
\caption{The structure of coverage-aware LSTM-based trajectory prediction.}
\label{fig:lstm}
\end{figure}

\section{LSTM-based Trajectory Prediction for Avatar Task Pre-migration Decisions}\label{lstm}
In this section, we first introduce the LSTM model and describe how we apply the LSTM to trajectory prediction. Then, we present our proposed LSTM-based trajectory prediction model to determine the expected workload of RSUs.
\subsection{Long Short-Term Memory Networks} 
LSTM is a variant of RNNs that effectively addresses the gradient explosion and vanishing gradient problems of RNNs and offers significant advantages in predicting long-term data series. LSTM uses a unique gate structure to selectively transfer and forget information. As shown in Fig.~\ref{fig:lstm}, each LSTM block consists of three gate structures, namely the forgetting gate $f_t$, the input gate $i_t$, and the output gate $o_t$. Similar to~\cite{yu2019review}, the three gate structures are defined as follows:
\begin{align}
f_{t} &= \sigma_{a}\left(W_{f} \cdot\left[h_{t-1}, x_{t}\right]+b_{f}\right), \\
i_{t} &= \sigma_{a}\left(W_{i} \cdot\left[h_{t-1}, x_{t}\right]+b_{i}\right), \\
\tilde{C}_{t} &= \tanh \left(W_{C} \cdot\left[h_{t-1}, x_{t}\right]+b_{C}\right), \\
C_{t} &= f_{t} * C_{t-1}+i_{t} * \tilde{C}_{t}, \\
o_{t} &= \sigma_{a}\left(W_{o} \cdot\left[h_{t-1}, x_{t}\right]+b_{o}\right), \\
h_{t} &= o_{t} * \tanh \left(C_{t}\right),
\end{align}
where $\sigma_{a}$ denotes the activation function, $W_f, W_i, W_C$, and $W_o$ represent the weight matrices of the input vector $x_t$ at time step $t$, $h_{t-1}$ is the state of the hidden layer at time step $t-1$ and $h_{t}$ is the state of the hidden layer at time step $t$. Additionally, $b_f, b_i, b_C$, and $b_o$ denote the bias vectors of the input vector $x_t$ at time step $t$, $*$ represents element-wise matrix multiplication, and $C_t$ represents the cell state.

The trajectory of a vehicle refers to a sequence of recorded positions over a period of time. A historical trajectory describes the vehicle's past positions, while a future trajectory indicates its prospective positions~\cite{ip2021vehiclelstm}.

To predict the future trajectories of intelligent vehicles, as shown in Fig.~\ref{fig:lstm}, we construct a five-layer LSTM model for trajectory prediction. In particular, the first layer captures the high-level features in the historical trajectory of the intelligent vehicle. The second layer is a dropout layer, which is used to prevent overfitting of the LSTM and improve its generalizability. The third layer further processes the features extracted from the previous layer to extract higher-level features that can capture more complex relationships of the intelligent vehicle's historical trajectory. Finally, the fourth and fifth layers are the fully connected layer and the ReLU layer, respectively, to output the future trajectory of the intelligent vehicle.

\begin{algorithm}[t]
\caption{Coverage-aware LSTM-based Trajectory Prediction.}\label{alg:lstm}
\begin{algorithmic} 
\STATE 
\STATE \textbf{Input:} The set of RSUs ${\{1,\dots,m,\dots,M\}}$ and the set of intelligent vehicles ${\{1,\dots, v,\dots,V\}}$ in the environment.\\
\STATE \textbf{Initialization:} Trajectory prediction model $J_{v}$.\\
Obtain location of RSU $P_{m}$, current location of intelligent vehicle $P_{v}(t)$, historical trajectories $\tau_{v}^{past}(t)$, and future trajectories $\tau_{v}^{fut}(t)$.\\
\STATE \textbf{Training Process}
\FOR{$e=0,1,\ldots,E-1$}
\STATE Input historical trajectories $\tau_{v}^{past}(t)$ to trajectory prediction model $J_{v}$, then obtain predicted trajectories $\tau_{v}^{pre}(t)$;\\
\STATE Calculate the MSE loss $\mathcal{L}_{v}^{train} = (\tau_v^{pre}(t)-\tau_v^{fut}(t))^2$ for each intelligent vehicle's trajectory;\\
\STATE Update the trajectory prediction model $J_{v}$ via back-propagation based on $\mathcal{L}_{v}^{train}$;\\
\ENDFOR
\STATE \textbf{Evaluating Process}
\STATE Input the historical trajectories $\tau_{v}^{past}(t')$ into the trained trajectory prediction model $J_{v}$, then obtain predicted trajectories $\tau_{v}^{pre}(t')$;\\
\STATE Calculate the MSE loss $\mathcal{L}_{v}^{test} = (\tau_v^{pre}(t')-\tau_v^{fut}(t'))^2$ for each intelligent vehicle's trajectory;\\
\FOR{$m=1,\ldots,M$}
\STATE Input predicted trajectories $\tau_{v}^{pre}(t')$ into RSU $m$ region selector to obtain the predicted traffic volume $z_m^{t'}$ in the RSU $m$ service region;\\
\STATE Calculate the predicted workload of RSU $m$: $L_m^{pre}(t')= L_m(t'-1) + \zeta z_m^{t'}$;\\
\ENDFOR
\STATE \textbf{Output:} Model $J_{v}$, MSE loss $\mathcal{L}_{v}^{test}$ of model $J_{v}$ and predicted workload of RSU $m$.
\end{algorithmic}
\label{alg1}
\end{algorithm}

\subsection{Coverage-aware LSTM-based Trajectory Prediction for Pre-migration Decision}
In AeVeM, the high mobility of intelligent vehicles is a significant factor that contributes to the dynamic workload of RSUs. Therefore, it is important to capture the movement patterns of intelligent vehicles to calculate the potential future workload of RSUs.

We first apply the trajectory prediction model mentioned earlier to predict the future trajectories of intelligent vehicles. As described in Algorithm~\ref{alg:lstm}, we first obtain the set of intelligent vehicles in the environment. Next, we obtain the historical trajectories of the intelligent vehicles $\tau_{v}^{past}(t)$ as well as the future trajectories $\tau_{v}^{fut}(t)$. During the training process, in each epoch, the intelligent vehicles' historical trajectories are fed into the trajectory prediction model, and the predicted trajectories are obtained. Subsequently, the future trajectories and predicted trajectories of each intelligent vehicle are used to evaluate the loss of the model. We use the Mean Square Error (MSE) to evaluate the loss of the model, which can be calculated as $\mathcal{L}_{v}^{train} = (\tau_v^{pre}(t)-\tau_v^{fut}(t))^2$. After that, the trajectory prediction models are updated based on $\mathcal{L}_{v}^{train}$ via back-propagation.

After the training of prediction model, in the evaluation process, the intelligent vehicles' historical trajectories $\tau_{v}^{past}(t')$ are fed into the trained trajectory prediction model and the predicted trajectories $\tau_{v}^{pre}(t')$ can be obtained. Similarly, the MSE is used to evaluate the performance of the trained model and apply in back-propagation of the training model, which can be calculated as $\mathcal{L}_{v}^{test} = (\tau_v^{pre}(t')-\tau_v^{fut}(t'))^2$.

Based on the identified future trajectories of intelligent vehicles, we propose a method to predict the future workload of RSUs. Specifically, we first obtain the set of RSUs ${\{1,\dots,m,\dots,M\}}$ in the environment and the location $P_{m}$ corresponding to each RSU. Next, as shown in Fig.~\ref{fig:lstm}, the predicted traffic data is transferred to the region selector of each RSU to obtain the predicted traffic $z_m^{t'}$ in the RSU $m$ service region. Finally, the predicted traffic volume $z_m^{t'}$ is fed into the RSU workload predictor, and the predicted workload of the RSU $m$ at time slot $t'$ can be calculated as 
\begin{equation}
L_m^{pre}(t')= L_m(t'-1) + \zeta z_m^{t'},\label{predload}
\end{equation}
where $\zeta$ is a coefficient that converts the predicted traffic volume to the predicted workload of the RSU.

The computation complexity of the proposed model evaluation process depends mainly on the complexity of the LSTM layer and the fully connected layer. Specifically, the computation complexity of the LSTM layer is $O(nd^2)$, where $n$ is the length of the input sequence and $d$ is the number of neurons. The computation complexity of the fully-connected layer is $O(n'd)$, where $n'$ is the length of the output sequence. Therefore, the total computation complexity is $O(jnd^2+n'd)$, where $j$ is the number of LSTM layers~\cite{freire2021experimental}.

\section{Multi-agent Deep Reinforcement Learning Scheme for Pre-migration Decisions}\label{alg}
\subsection{POMDP for Avatar Task Pre-migration}
In AeVeM, the key factors affecting the pre-migration decision of intelligent vehicle $v$ at time slot $t$ include the mobility of the intelligent vehicle, the dense of RSUs, the workload of RSUs, the avatar tasks inputs from other intelligent vehicles, and pre-migration decisions of intelligent vehicles. Ideally, when intelligent vehicle $v$ has complete information related to the decision process, it can make the optimal pre-migration decision for the avatar task. However, it is impractical for intelligent vehicles to obtain the future workload of RSUs before making pre-migration decisions for avatar tasks. At each time slot, the workload of RSUs is affected by the avatar tasks of other intelligent vehicles. To enable intelligent vehicles to make optimal decisions based on partially observable information when they cannot accurately observe the state of their environment, we transform the avatar task migration problem into a POMDP, which is defined as follows. 

\begin{figure*}[t]  	
\centerline{\includegraphics[width=1\linewidth]{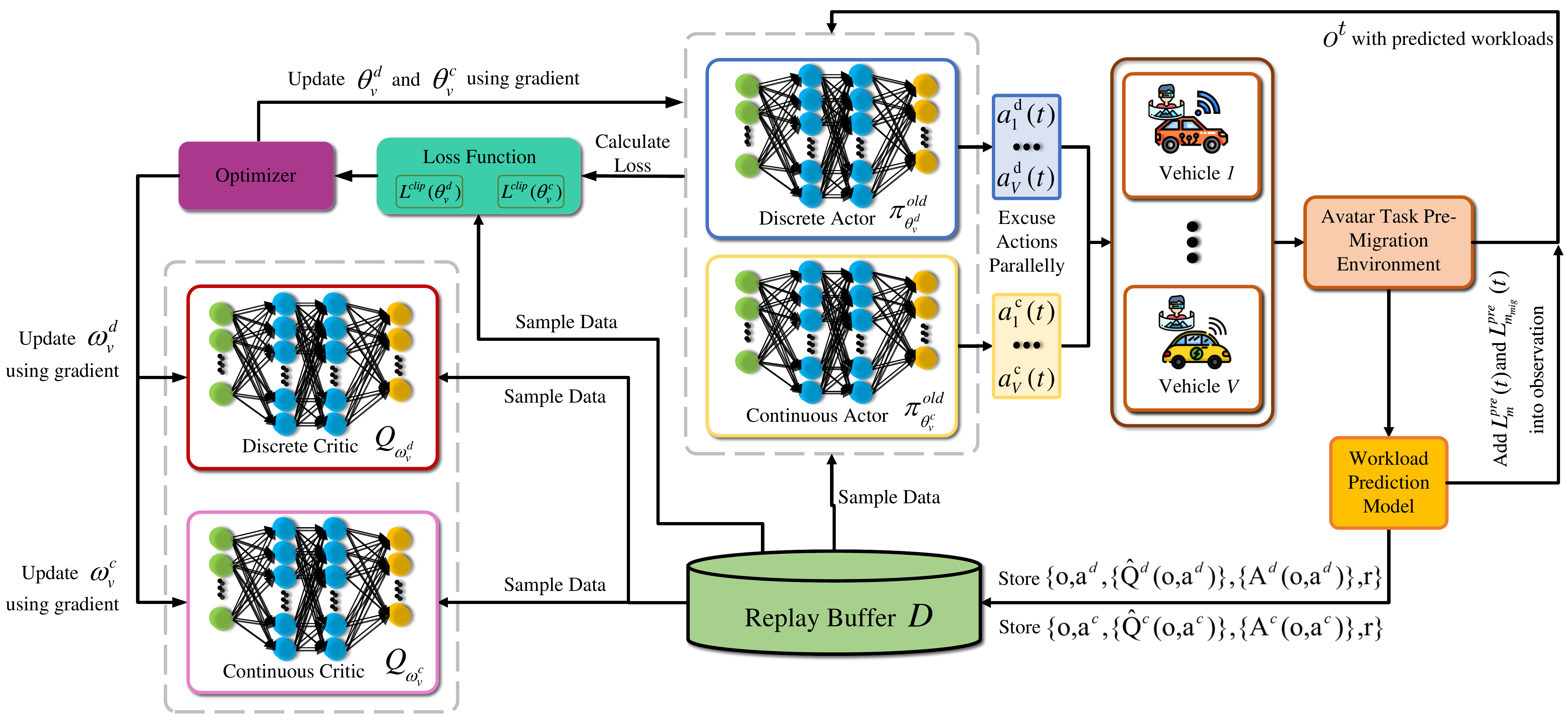}}  	
\caption{Architecture of the Hybrid-MAPPO algorithm for avatar tasks pre-migration.} \label{fig:hmappo}  
\end{figure*}

\begin{itemize}
    \item \textbf{Observation: } In AeVeM, intelligent vehicles are able to observe real-time information about RSUs and intelligent vehicles in the avatar task migration environment. The ensemble of observations for all intelligent vehicles in the environment is defined as $\mathcal{O}$, which can be expressed as $\mathcal{O} = \{ o_1, \dots,o_v, \dots, o_V\}$. For intelligent vehicle $v$, its observation at time slot $t$ is defined as
    \begin{equation}
        o_v^t=[P_v^t,L_m^{pre}(t),L_{m_{p}}^{pre}(t),T^{total}(t)],\label{obs}
    \end{equation}
    where $P_v^t$ represents the current position of the intelligent vehicle, $L_m^{pre}(t)$ is the predicted workload of RSU $m$ at time slot $t$, $L_{m_{p}}^{pre}(t)$ is the expected workload of RSU $m_{p}$ at time slot $t$, and $T^{sum}_v(t)$ is the total avatar services latency of intelligent vehicle $v$ at time slot $t$. The expected workload of RSUs can be obtained from the analysis of future trajectories of intelligent vehicles according to the proposed LSTM-based trajectory prediction model.
    \item \textbf{Action:} At each time slot, intelligent vehicle $v$ can simultaneously perform actions with two action spaces, i.e., $K^{rd}_v(t)$ or $K^{ud}_v(t)$ with discrete action space and $K^p_v(t)$ with continuous action space. To simplify the representation, the aggregated actions of intelligent vehicle $v$ at time slot $t$ can be expressed as 
    \begin{equation}
    a_v^t = \{ a_v^d(t),a_v^c(t)\},\label{action}
    \end{equation}
    where $a_v^d(t)$ denotes the discrete decision as $K^{rd}_v(t)$ in remote areas, $K^{ud}_v(t)$ in urban areas while $a_v^c(t)$ denotes the continuous decision $K^p_v(t)$. Furthermore, we define the joint action of all intelligent vehicles at time slot $t$ as $a$.

    \item \textbf{Reward:} At each time slot, each intelligent vehicle gets an observation $ o_v^t$ and performs action $a_v^t$ in the environment according to the migration policy. The reward returned by the environment during the interaction is defined as
    \begin{equation}
    r_v^t (o_v^t, a_v^t)= -T^{sum}_v(t), \label{eqrew}
    \end{equation}
    where $T^{sum}_v(t)$ is the total avatar service latency of intelligent vehicle $v$ at time slot $t$.
\end{itemize}

\begin{minipage}{0.45\textwidth}
\begin{algorithm}[H]
\caption{Hybrid-MAPPO Algorithm for Avatar Task Pre-migration}
\label{alg:algorithm}
\begin{algorithmic}[1] 
\STATE Initialize discrete actor $\pi_{\theta_v^d}$, $\pi_{\theta_v^d}^{old}$, discrete critic $Q_{\omega_v^d}$, $Q_{\bar{\omega}_v^d}$, continuous actor $\pi_{\theta_v^c}$, $\pi_{\theta_v^c}^{old}$, continuous critic $Q_{\omega_v^c}$, $Q_{{\overline{\omega}}_v^c}$;
\STATE Initialize environment $Env$, Initialize replay buffer $D$;
\FOR {Episode $1,2,\dots,E$}
\FOR{Time slot $t=1,2,\dots,T$}
\STATE Add predicted workload into the observation of each agent $v$;\\
\STATE Each agent $v$ gets $a_v^t = \{ a_v^d(t),a_v^c(t)\}$ according to $o_v^t$ by discrete actor $\pi_{\theta_v^d}^{old}$ and continuous actor $\pi_{\theta_v^c}^{old}$ respectively;\\
\STATE Get next state $o^{t+1}$ and reward $r^t$;
\ENDFOR
\STATE Each agent $v$ gets a trajectory $\tau_v=\{o_v^t,a_v^t,r_v^t\}_{t=1}^{T}$;
\STATE Compute ${\{\hat{Q}_v^d[o^t,a^d(t)]\}}_{t=1}^{T}$ according to eq.(\ref{hatqd});
\STATE Compute ${\{\hat{Q}_v^c[o^t,a^c(t)]\}}_{t=1}^{T}$ according to eq.(\ref{hatqc});
\STATE Compute ${\{A_v^d[o^t,a^d(t)]\}}_{t=1}^{T}$ according to eq.(\ref{advd});
\STATE Compute ${\{A_v^c[o^t,a^c(t)]\}}_{t=1}^{T}$ according to eq.(\ref{advc});
\STATE Store ${\{o,a^d,\{\hat{Q}^d(o,a^d)\},\{A^d(o,a^d)\},r\}}$ and ${\{o,a^c,\{\hat{Q}^c(o,a^c)\},\{A^c(o,a^c)\},r\}}$ into replay buffer $D$;

\FOR{epoch k=1,2,...,K}
\STATE Shuffle the data in the buffer $D$
\FOR{j=0,1,2,...,{$\frac{T}{G}$}-1}
\STATE Sample $G$ mini-batch of data from buffer $D$ 
\FOR{v=1,2,...,V}
\STATE $\Delta\theta_v^d=\frac{1}{G}\sum_{i=1}^G\{{\nabla}_{\theta_v^d}F(\beta_{\theta_v^d}^i,A_v^d[o_i,a_i])\}$
\STATE $\Delta\theta_v^c=\frac{1}{G}\sum_{i=1}^G\{{\nabla}_{\theta_v^c}F(\beta_{\theta_v^c}^i,A_v^c[o_i,a_i])\}$
\STATE $\Delta\omega_v^d=\frac{1}{G}\sum_{i=1}^G\{{\nabla}_{\omega_v^d}[\hat{Q}_v(o_i,a_i)-Q_{\omega_v^d}(o_i,a_i)]^2\}$
\STATE $\Delta\omega_v^c=\frac{1}{G}\sum_{i=1}^G\{{\nabla}_{\omega_v^c}[\hat{Q}_v(o_i,a_i)-Q_{\omega_v^c}(o_i,a_i)]^2\}$
\ENDFOR
\STATE Update $\theta_v^d$, $\theta_v^c$ using $\Delta\theta_v^d$, $\Delta\theta_v^c$;
\STATE Update $\omega_v^d$, $\omega_v^c$ using $\Delta\omega_v^d$, $\Delta\omega_v^c$;
\ENDFOR
\ENDFOR
\STATE Update $\theta_v^{d(old)}\xleftarrow{} \theta_v^d $, $\theta_v^{c(old)}\xleftarrow{} \theta_v^c $;
\STATE Update $\overline{\omega}_v^d\xleftarrow{}\omega_v^d$, $\overline{\omega}_v^c\xleftarrow{}\omega_v^c$.
\ENDFOR
\end{algorithmic}
\end{algorithm}
\vspace{0.05cm}
\end{minipage}

\subsection{Hybrid-MAPPO Algorithm Design}
Based on the collaboration of multiple learning agents, the Proximal Policy Optimization (PPO) algorithm can be extended to the MAPPO algorithm. Unlike the decentralized training and decentralized execution framework used by the independent PPO algorithm, the MAPPO algorithm employs centralized training and decentralized execution. Specifically, each agent in MAPPO has an independent policy network and shares the common value network. This scheme allows all agents to share the global valuation. This improves collaboration between agents and enables better performance in a multi-agent collaborative environment.

Since the action space of the proposed problem contains both discrete and continuous actions, the vanilla MAPPO algorithm cannot be applied~\cite{qiu2023hierarchical}. In this paper, we extend the MAPPO algorithm to a MADRL algorithm that can perform both discrete and continuous actions simultaneously, i.e., Hybrid-MAPPO. Specifically, we design two sets of strategies, namely, discrete and continuous strategies. For each learning agent $v$, its discrete action strategy and continuous action policy are defined as $\pi_{\theta_v^d}[a_v^d(t)|o_v^t]$ and $\pi_{\theta_v^c}[a_v^c(t)|o_v^t]$, respectively. 

The discrete and continuous strategies can be updated separately. To avoid instabilities caused by excessive differences in policy network updates, the discrete and continuous strategies minimize their respective clipping surrogate targets to constrain policy updates, which can be defined as
\begin{equation}
L^{clip}(\theta_v^d) = \mathbb{E}\{F(\beta_{\theta_v^d}^i,A_v^d[o_i,a_i])\},\label{ld}
\end{equation}
\begin{equation}
L^{clip}(\theta_v^c) = \mathbb{E}\{F(\beta_{\theta_v^c}^i,A_v^c[o_i,a_i])\},\label{cd}
\end{equation}
\begin{equation}
F(\beta_{\theta_v^d}^i,A_v^d[o_i,a_i])=min[\beta_{\theta_v^d}^t A_v^d,{f}_{clip}[\beta_{\theta_v^d}^t,\epsilon] A_v^d],\label{fund}
\end{equation}
\begin{equation}
F(\beta_{\theta_v^c}^i,A_v^c[o_i,a_i])=min[\beta_{\theta_v^c}^t A_v^c,{f}_{clip}[\beta_{\theta_v^c}^t,\epsilon] A_v^c],\label{fund}
\end{equation}
where ${f}_{clip}[\beta_{\theta_v^d}^t,\epsilon] A_v^d]$ and ${f}_{clip}[\beta_{\theta_v^c}^t,\epsilon] A_v^c]$ are the clip functions, $\epsilon\in[0,1]$ is the clipping parameter, $\beta_{\theta_v^d}^t$ and $\beta_{\theta_v^c}^t$ denote the ratio  between the new policies and the old policies, which are defined as
\begin{equation}
\beta_{\theta_v^d}^t=\frac{\pi_{\theta_v^d} [a_v^d(t)|o_v^t]}{\pi_{\theta_v^d}^{old} [a_v^d(t)|o_v^t]},\label{db} 
\end{equation}
\begin{equation}
\beta_{\theta_v^c}^t=\frac{\pi_{\theta_v^c} [a_v^c(t)|o_v^t]}{\pi_{\theta_v^c}^{old} [a_v^c(t)|o_v^t]},\label{db} 
\end{equation}
where $\pi_{\theta_v^d}^{old} [a_v^d(t)|o_v^t]$ and $\pi_{\theta_v^c}^{old} [a_v^c(t)|o_v^t]$ are the old policies. The joint advantage function $A_v^d[o^t,a^d(t)]$ and $A_v^c[o^t,a^c(t)]$ are used to measure the advantage of action $a^d(t)$ and $a^c(t)$ over the average action in state $o^t$ \cite{DelinGuo2020MAPPO}, which can be calculated as
\begin{equation}
A_v^d[o^t,a^d(t)]=\hat{Q}_v^d[o^t, a^d(t)]-b[o^t, a^d_{-v}(t)],\label{advd} 
\end{equation}

\begin{equation}
A_v^c[o^t,a^c(t)]=\hat{Q}_v^c[o^t, a^c(t)]-b[o^t, a^c_{-v}(t)],\label{advc} 
\end{equation}
where $\hat{Q}_{v}[o^t, a^d(t)]$ and $\hat{Q}_{v}[o^t, a^c(t)]$ are the estimates of the action-value function, which can be calculated as 
\begin{equation}
\begin{aligned}
\hat{Q}_v^d[o^t, a^d(t)]&=Q_{\bar{\omega}_v^d}[o^t, a^d(t)] \\
& \quad +\delta_{t}+(\gamma \lambda) \delta_{t+1}+\cdots+(\gamma \lambda)^{T} \delta_{T},\label{hatqd}
\end{aligned}
\end{equation}

\begin{equation}
\begin{aligned}
\hat{Q}v^c[o^t, a^c(t)] &= Q{\bar{\omega}v^c}[o^t, a^c(t)] \\
& \quad +\delta{t}+(\gamma \lambda) \delta_{t+1}+\cdots+(\gamma \lambda)^{T} \delta_{T},\label{hatqc}
\end{aligned}
\end{equation}
where $Q_{\bar{\omega}_v^d}[o^t, a^d(t)]$ and $Q_{\bar{\omega}_v^c}[o^t, a^c(t)]$ are the centralized critics, $\delta_{t}$ is the TD error, $\gamma$ is the discount factor, and $\lambda$ is the decay factor in TD error. $b[o^t, a^d_{-v}(t)]$ and $b[o^t, a^c_{-v}(t)]$ are counterfactual baselines, where $a^d_{-v}(t)$ and $a^c_{-v}(t)$ are the joint action of agents other than agent $v$.

Algorithm \ref{alg:algorithm} lists the pseudo-code of the proposed Hybrid-MAPPO algorithm. First, we initialize discrete actor $\pi_{\theta_v^d}$, $\pi_{\theta_v^d}^{old}$, discrete critic $Q_{\omega_v^d}$, $Q_{\bar{\omega}_v^d}$, continuous actor $\pi_{\theta_v^c}$, $\pi_{\theta_v^c}^{old}$ and continuous critic $Q_{\omega_v^c}$, $Q_{{\overline{\omega}}_v^c}$. Next, we initialize the avatar task pre-migration environment $Env$ and replay buffer $D$.

As shown in Fig.~\ref{fig:hmappo}, during the training process, each agent's discrete actor and continuous actor output discrete action $a_v^d(t)$ and continuous action $a_v^c(t)$ simultaneously based on their observations $o^t_v$, respectively. Then, agents receive the next state $o^{t+1}$ and reward $r^t$ returned by the environment. After that, each agent $v$ can get the trajectory $\tau_u$ and calculate ${\{\hat{Q}_v^d[o^t,a^d(t)]\}}_{t=1}^{T}$, ${\{\hat{Q}_v^c[o^t,a^c(t)]\}}_{t=1}^{T}$, ${\{A_v^d[o^t,a^d(t)]\}}_{t=1}^{T}$ and ${\{A_v^c[o^t,a^c(t)]\}}_{t=1}^{T}$ respectively. The experiences ${\{o,a^d,\{\hat{Q}^d(o,a^d)\},\{A^d(o,a^d)\},r\}}$ and ${\{o,a^c,\{\hat{Q}^c(o,a^c)\},\{A^c(o,a)\},r\}}$ are stored in the replay buffer $D$ after the transition of environment is done. In each training epoch $k$, the order of the data can be disrupted to break the correlation between samples, thus improving stability. Then, $\theta_v^d$, $\theta_v^c$, $\omega_v^d$ and $\omega_v^c$ are updated using gradient $\Delta\theta_v^d$, $\Delta\theta_v^c$, $\Delta\omega_v^d$ and $\Delta\omega_v^c$ with $G$ mini-batch of data sampled from the replay buffer, respectively. After $K$ training epochs, $\theta_v^{d(old)}$, $\theta_v^{c(old)}$, $\bar{\omega}_v^d$ and $\bar{\omega}_v^c$ of each agent are updated to the new $\theta_v^{d}$, $\theta_v^{c}$, ${\omega}_v^d$ and ${\omega}_v^c$.

Combined with the prediction model, the total computation complexity of the Hybrid-MAPPO algorithm with predicted workload can be calculated as $O[ETP(jnd^2+n'd)V(S' + \frac{K}{D})]$, where $S'$ represents the input size of actor-network for all agents, $P$ represents the kind of different actions for all agents~\cite{gong2022computation}.

\section{Numerical Results}\label{exp}
In this section, we first present the parameter settings for the experiments with the avatar task scenario before migration. Then, we present the experimental results of our proposed coverage-aware LSTM trajectory prediction. Next, we perform the convergence analysis of our proposed Hybrid-MAPPO algorithm and compare it with other baseline algorithms. Finally, we evaluate the performance of the Hybrid-MAPPO algorithm and other baseline algorithms in different scenarios.

\subsection{Parameter Settings}

Initially, we present the parameter settings for the avatar task pre-migration scenario experiments. An urban scenario with 10 intelligent vehicles and 6 RSUs is being examined evenly distributed on both sides of the road. Each RSU has a wireless bandwidth between 200 MHz and 600 MHz, a migration bandwidth between 500 Mbps and 900 Mbps, and GPU computing resources between 10 GHz and 30 GHz. The cloud servers have a computing capability in the range of 50 GHz to 100 GHz, and the required GPU cycles per unit data of the vehicle are set to 0.5 Gcycles/MB. For the intelligent vehicles, the avatar task input size ranges from 12 MB to 20 MB, and the avatar task size ranges from 50 MB to 250 MB. The channel gain of the intelligent vehicle is set at 4.11, and the carrier frequency is set at 2.0 GHz.

In the coverage-aware LSTM Trajectory prediction, we set the historical trajectory acquisition steps of the intelligent vehicle to 12 and the future trajectory prediction steps to 1. For the LSTM trajectory prediction model, we set the number of cells in each LSTM layer to 256, and the dropout rate of the dropout layer to 0.05. Additionally, we set the batch size to 40 and the number of epochs to 500. We use Keras to build the LSTM trajectory prediction model~\cite{Keras}.

In the Hybrid-MAPPO algorithm, the learning rate is set to $1*10^{-3}$ for optimization. The discount factor, denoted as $\gamma$, is assigned a value of 0.95, while the clipping parameter $\epsilon$ is set to 0.2. Additionally, the buffer size $D$ is defined as 20000 to facilitate efficient training.
\subsection{Coverage-aware LSTM Trajectory Prediction}
\begin{figure}[t]
\centerline{\includegraphics[width=0.45\textwidth]{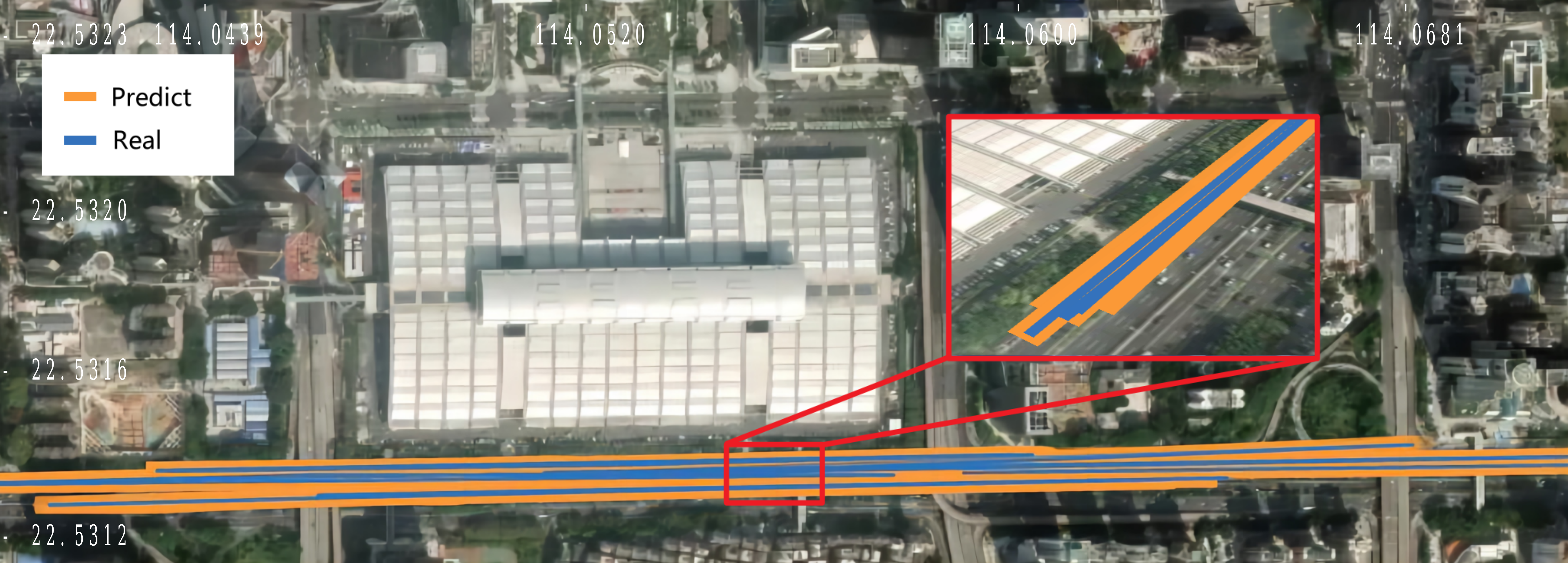}}
\caption{The comparison of real trajectories and predicted trajectories.}
\label{fig:lstmresult}
\end{figure}

\begin{table}[t]
\centering
\caption{Testing results of trajectory prediction.}
\begin{tabular}{|l|l|l|l|} 
\hline
Metric & Value & Metric & Value\\
\hline
Test MSE & $6.3*10^{-5}$ & Test MAE & $6.032*10^{-3}$ \\
\hline
Test $r^2$ & $9.987*10^{-1}$ & Test MedAE & $4.949*10^{-3}$ \\
\hline
\end{tabular}
\label{tab:results}
\end{table}

To validate the effectiveness of our proposed trajectory prediction model, the dataset in \cite{wang2019dataset} is leveraged for trajectory prediction and visualizes the output future trajectories, which contains mobility traces of 664 vehicles in the Chinese city of Shenzhen over one day. In the experiment, we observe 10 vehicles running in an urban area, whose latitude and longitude are from 22.5312 to 23.5323 and from 114.0439 to 114.0681, respectively. As shown in Fig.~\ref{fig:lstmresult}, the blue trajectories represent the real trajectories of the intelligent vehicles, while the orange trajectories represent the predicted trajectories of the intelligent vehicles using LSTM trajectory prediction model. This figure also presents the predicted trajectories that are highly fitted to the real trajectories, which demonstrates the effectiveness of our proposed trajectory prediction model.

To further demonstrate the effectiveness of our proposed trajectory prediction model, four commonly-used metrics are employed to evaluate the performance of the tested model~\cite{zhou2023improved,sareen2023imputation}. First, the MSE is a metric to assess the average deviation between the predicted and true results of the model. Second, the R2 score ($r^2$) is a metric to assess how well the model fits the data. The value of $r^2$ ranges from 0 to 1, and the closer it is to 1, the better the model fit. Third, the Mean Absolute Error (MAE) assesses the mean absolute deviation between the predicted and true results of the model. Finally, the Median Absolute Error (MedAE) evaluates the median absolute deviation between the predicted and true results of the model.

As shown in Table~\ref{tab:results}, our test model is evaluated using four performance metrics. First, the MSE of our test model is 0.000063, indicating that the average deviation between our model's predicted results and the true results is very small. Second, the $r^2$ of our test model was 0.998746, which is very close to 1, indicating a strong fit between our model and the data. Third, the MAE of our test model is 0.006032, indicating a very small average absolute deviation between our model's predicted results and the true results. Finally, the MedAE of our test model is 0.004949, further indicating the small deviation between our model's predicted results and the true results. Collectively, these results demonstrate the effectiveness of our proposed trajectory prediction model.

\subsection{Convergence Analysis}
To demonstrate the effectiveness of our proposed Hybrid-MAPPO algorithm and to verify the impact of Coverage-aware LSTM Trajectory Prediction on the algorithm's performance, we design several baselines for comparison, including \textit{Hybrid-MAPPO w.o. Prediction}, \textit{MAPPO-DDPG with Prediction}, \textit{MAPPO-DDPG w.o. Prediction}, \textit{Greedy}, \textit{Random}, \textit{Full Pre-migration (FPM)}, and \textit{No Pre-migration (NPM)}, where \textit{w.o. Prediction} denotes the scheme that the observation of agents without RSU workload prediction. The Multi-Agent Deep Deterministic Policy Gradient (MADDPG) algorithm is widely adopted as a representative of multi-agent reinforcement learning algorithms that employ continuous actions. For fairness during comparison, the MADDPG algorithm also adopts a centralized training and distributed execution framework. Therefore, we compare our algorithm with the MAPPO-DDPG algorithm consisting of discrete-action MAPPO and continuous-action MADDPG.
\begin{figure}[t]
\vspace{-0.3cm}
\centerline{\includegraphics[width=0.4\textwidth]{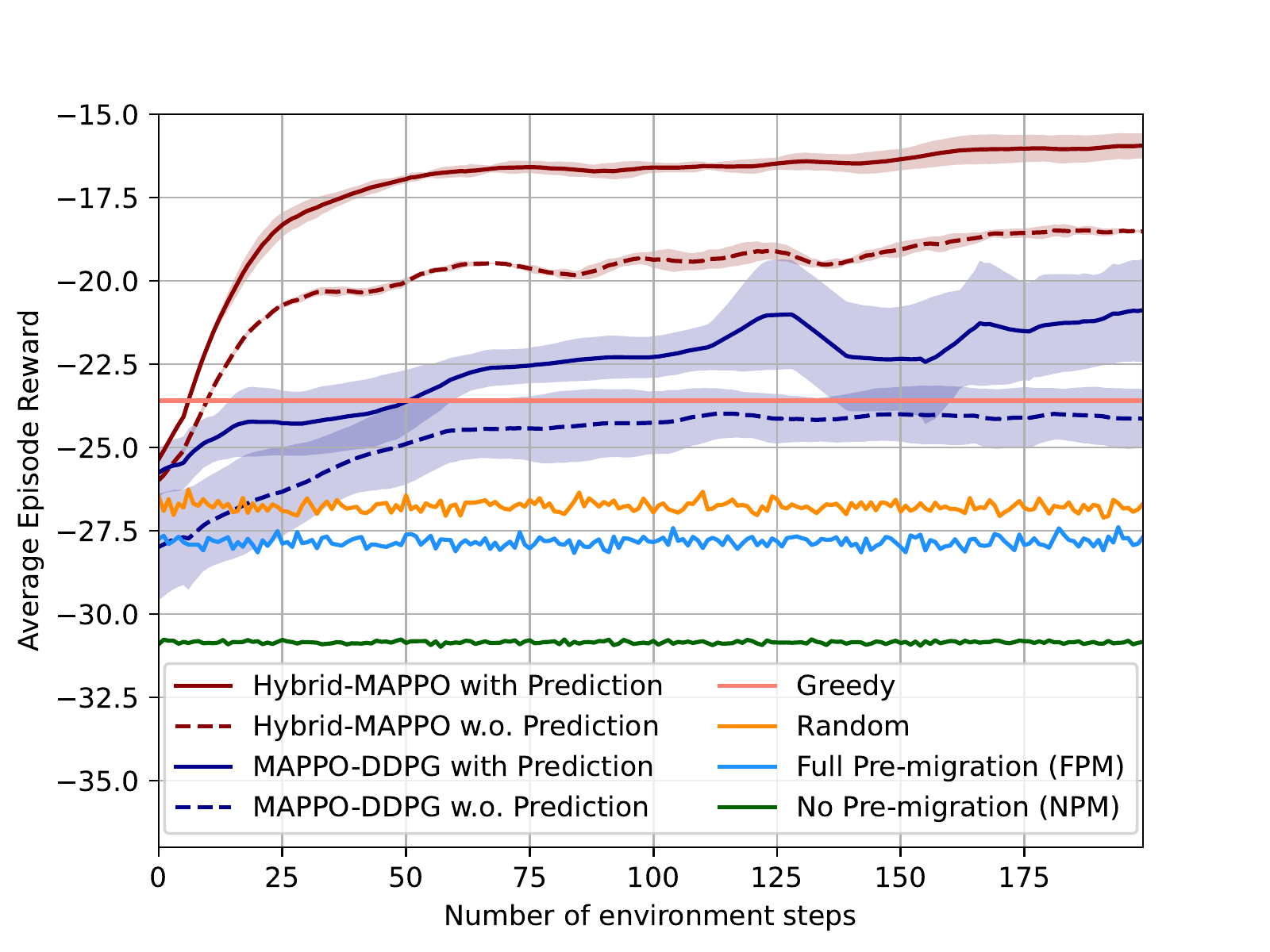}}
\caption{Average episode reward versus number of environment steps.}
\label{fig:converge}
\end{figure}

As shown in Fig.~\ref{fig:converge}, our proposed scheme significantly outperforms the baseline scheme. Specifically, our proposed scheme outperforms \textit{Hybrid-MAPPO w.o. Prediction}, \textit{MAPPO-DDPG with Prediction}, \textit{MAPPO-DDPG w.o. Prediction}, \textit{Greedy}, \textit{Random}, \textit{FPM}, and \textit{NPM} by 13\%, 24\%, 33\%, 32\%, 40\%, 42\%, and 48\%, respectively. As illustrated by the curves in the figure, the average reward obtained by the scenarios with RSU workload prediction is about 11\% higher than that of the scheme without prediction. Additionally, the convergence speed of the scheme with RSU workload prediction is also higher than that of the scheme without it, which sufficiently demonstrates the effectiveness of incorporating RSU workload prediction in improving the performance of the algorithm. In Fig.~\ref{fig:converge}, the performance of the \textit{Greedy} scheme is worse than that of the reinforcement learning-based scheme. The reason is that each intelligent vehicle's action has a negative impact on the environment in the next time slot, while the \textit{Greedy} scheme considers only maximizing the current reward and ignores the long-term reward. Besides, the \textit{FPM} scheme and \textit{NPM} scheme perform worse than the random scheme, as always Pre-migrate all the avatar tasks add unnecessary migration latency, while No Pre-migrate avatar tasks tend to overload the RSUs, which leads to higher processing latency.

\begin{figure}[t]
\vspace{-0.3cm}
\centerline{\includegraphics[width=0.4\textwidth]{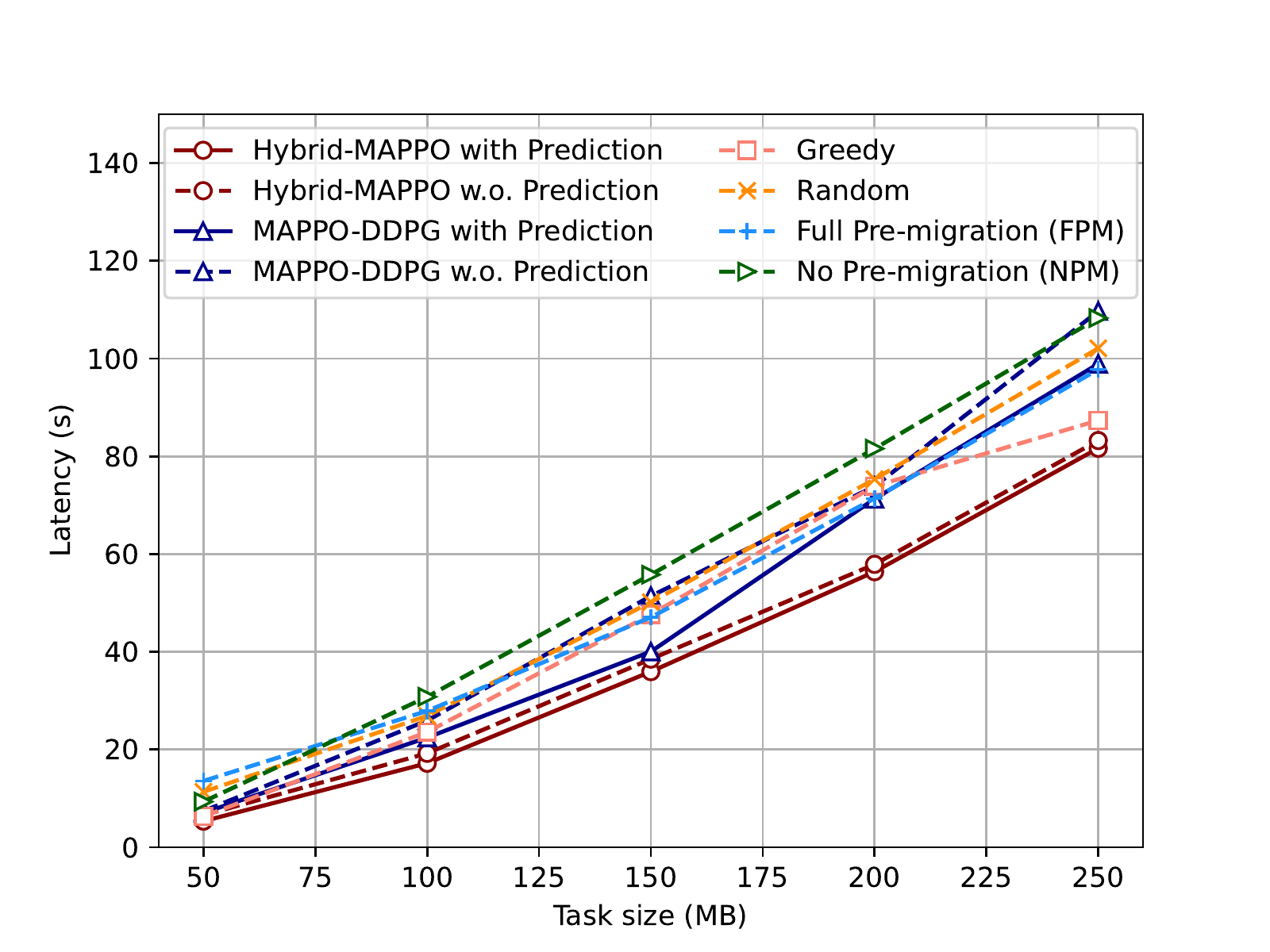}}
\caption{Average system latency versus different task sizes.}
\label{fig:size}
\end{figure}

\begin{figure}[t]
 \vspace{-0.3cm}
\centerline{\includegraphics[width=0.4\textwidth]{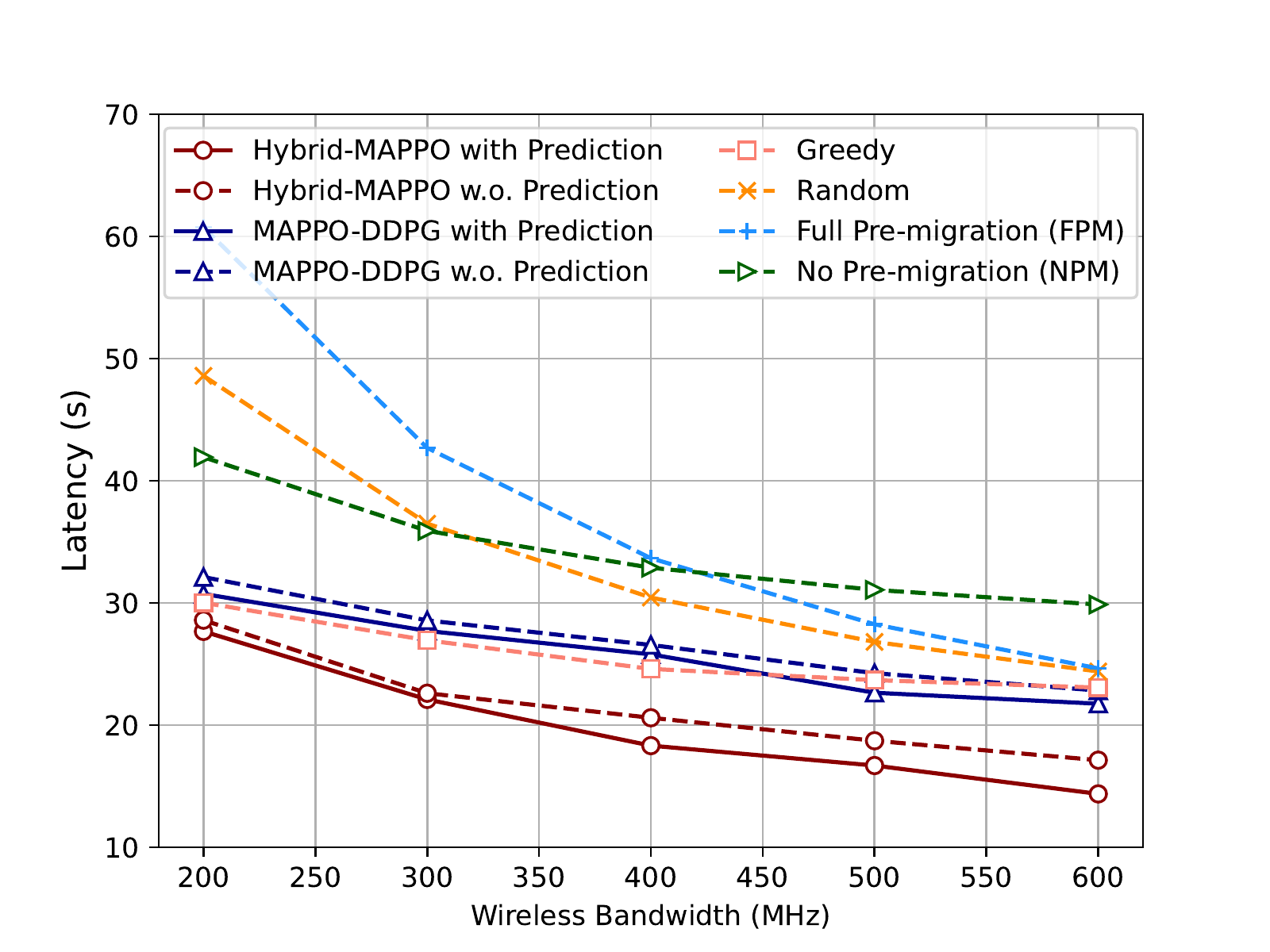}}
\caption{Average system latency versus different task migration bandwidth.}
\label{fig:wireless}
\end{figure}

\begin{figure}[t]
 \vspace{-0.3cm}
\centerline{\includegraphics[width=0.4\textwidth]{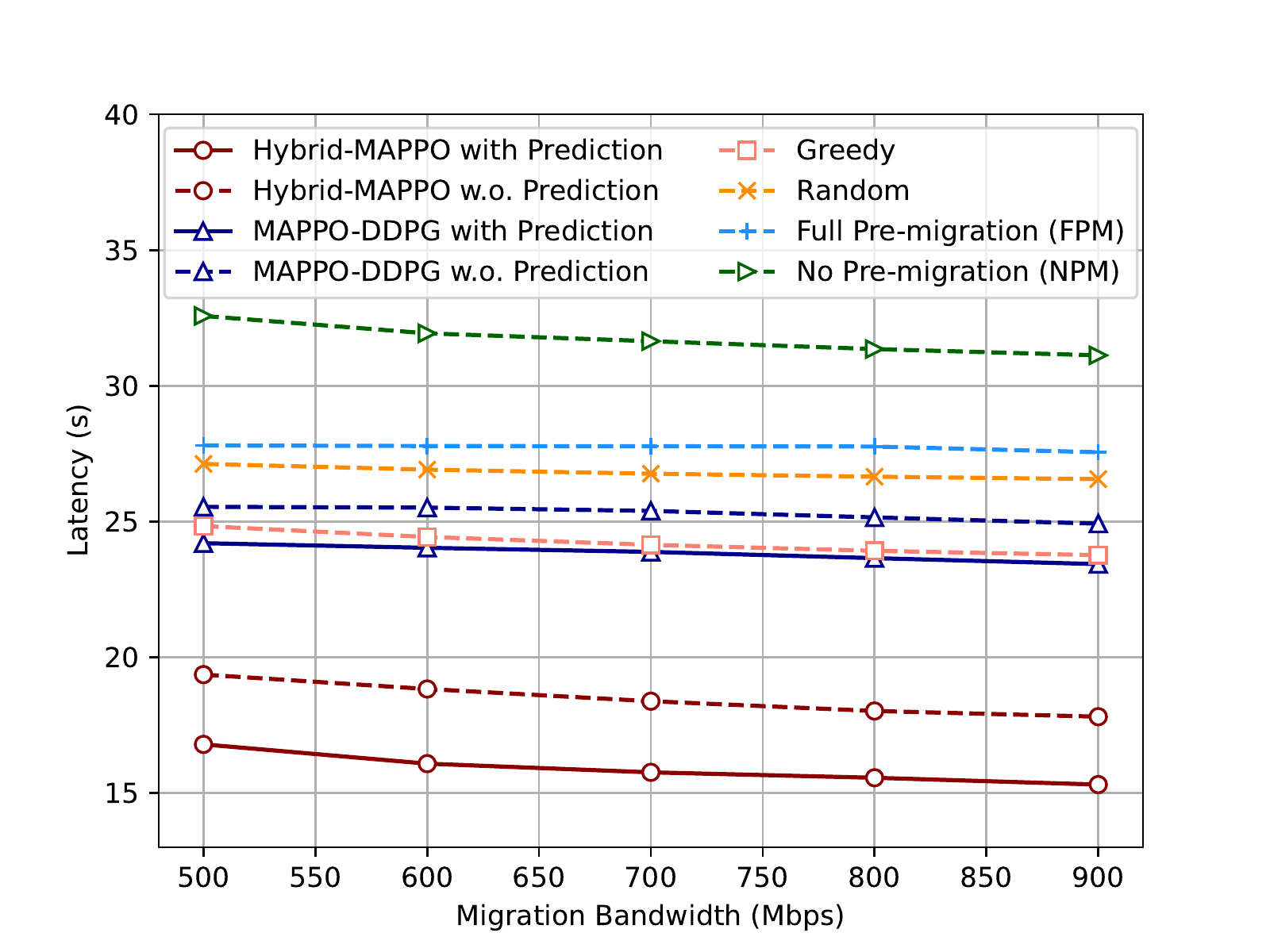}}
\caption{Average system latency versus different task migration bandwidth.}
\label{fig:migration}
\end{figure}

\begin{figure}[t]
\vspace{-0.3cm}
\centerline{\includegraphics[width=0.4\textwidth]{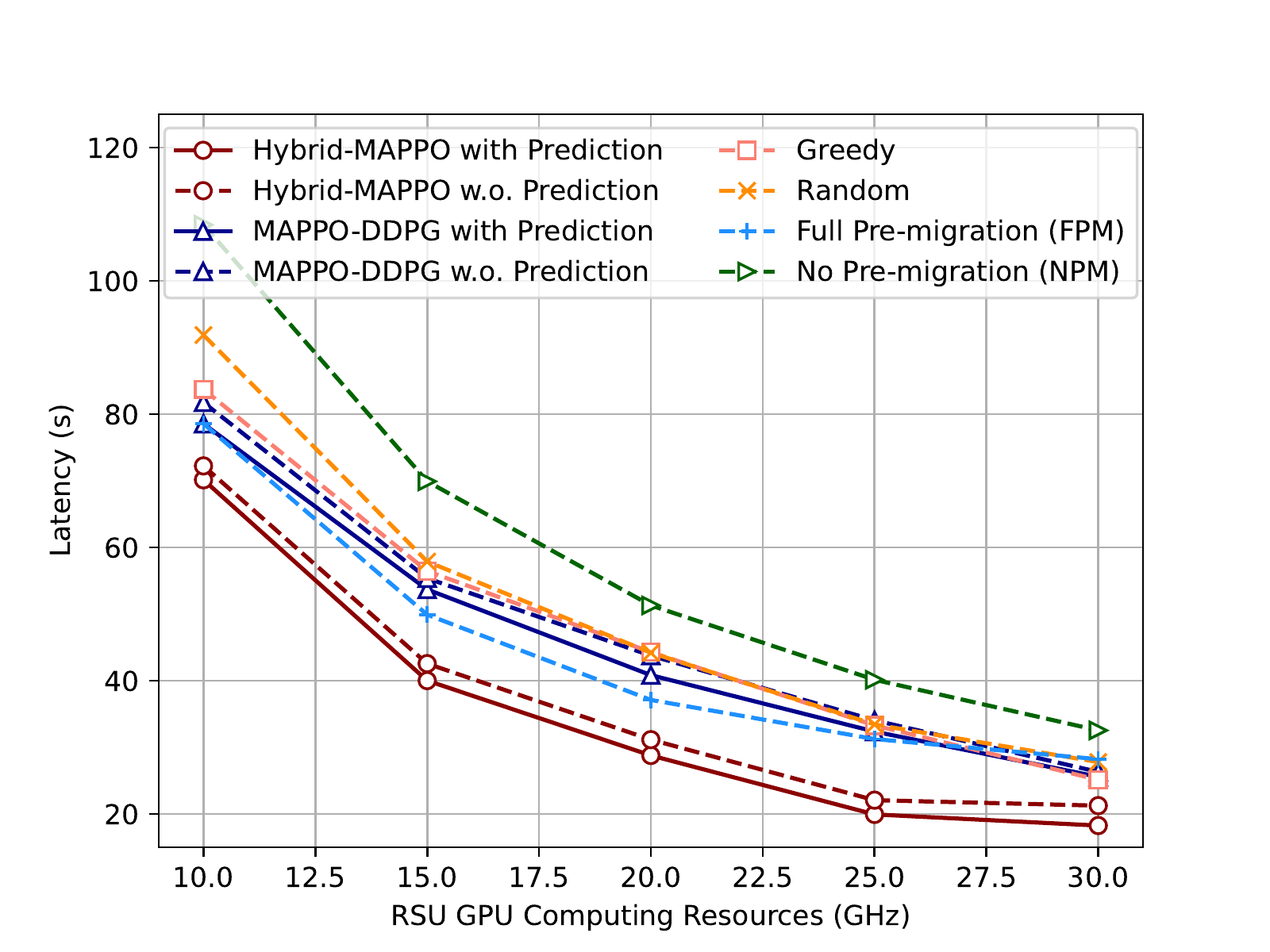}}
\caption{Average system latency versus different RSUs' computing capabilities.}
\label{fig:edge}
\end{figure}

\begin{figure}[t]
\vspace{-0.3cm}
\centerline{\includegraphics[width=0.4\textwidth]{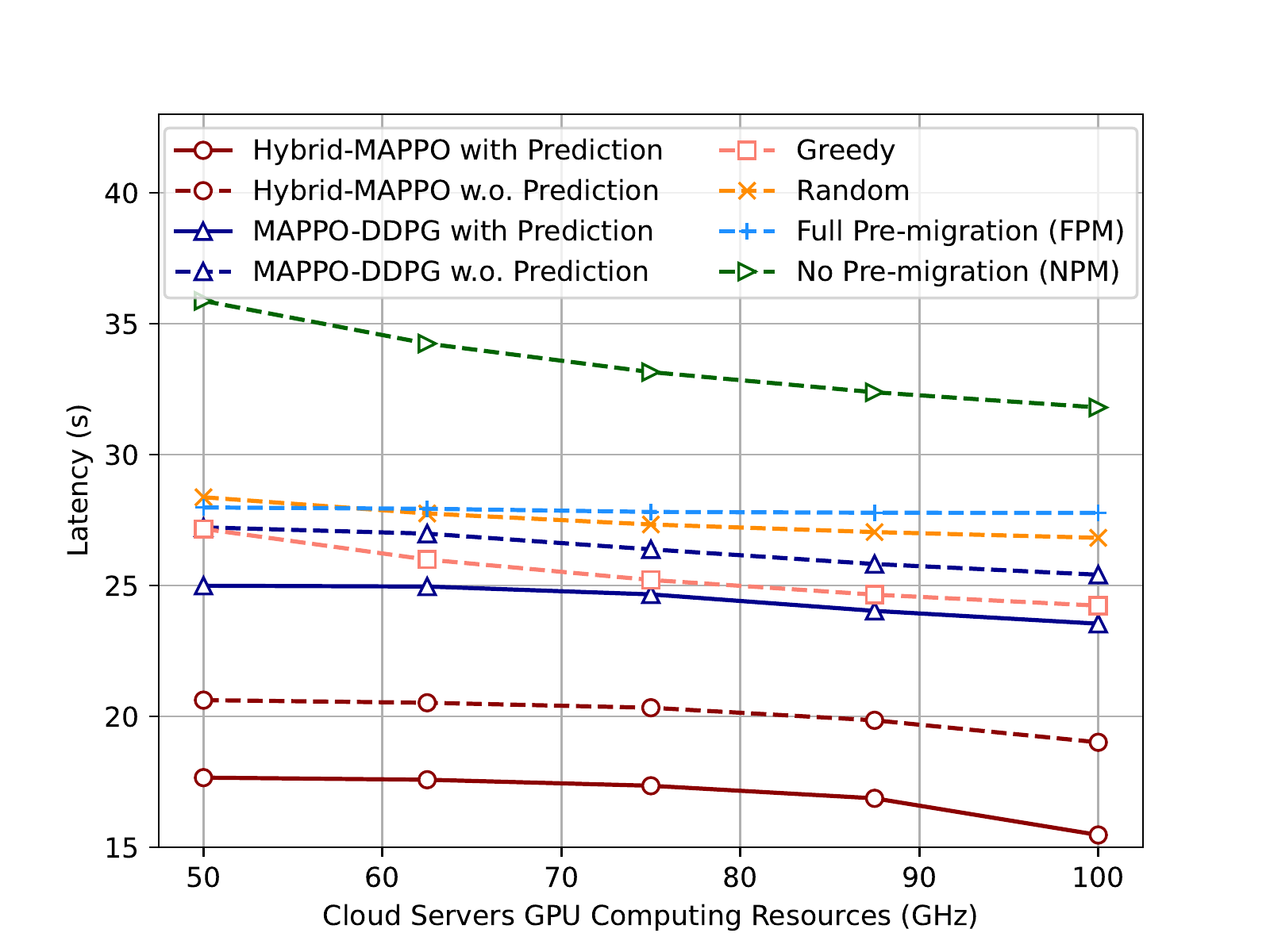}}
\caption{Average system latency versus different cloud computing capabilities.}
\label{fig:cloud}
\end{figure}

\subsection{Performance on Avatar Task Pre-migration}
To demonstrate the robustness of our scheme under different system settings, we used several different sets of environmental parameters to validate the proposed scheme.

First, we examine the impact of different avatar task sizes on system latency and evaluate the effectiveness of the proposed scheme in reducing system latency across different scenarios. Fig.~\ref{fig:size} examines the impact of different avatar task sizes on system latency under various scenarios. For avatar task sizes of 50 MB, 100 MB, 150 MB, 200 MB, and 250 MB, respectively, our proposed scheme achieves the lowest system latency compared to the baseline schemes, which proves that our scheme is effective in reducing system latency across different scenarios. As observed from the curves in the figure, the system latency increases as the avatar task size increases. The reason is that a larger avatar task not only increases the time required for RSUs to compute the task but also leads to an increase in the waiting time for other avatar tasks to be processed. Furthermore, the schemes with workload prediction perform better than the ones without workload prediction by 20\%, highlighting the importance of predicting the future workload of the RSU for making reasonable pre-migration decisions.

Then, we discuss the effect of different wireless bandwidths on system latency for various schemes and show that the proposed scheme achieves the lowest system latency compared to the baseline schemes for different wireless bandwidths. Fig.~\ref{fig:wireless} shows the effect of different wireless bandwidths on system latency for various schemes. For wireless bandwidths of 100 MHz, 200 MHz, 300 MHz, 400 MHz, and 500 MHz, our proposed scheme achieves the lowest system latency compared to the baseline schemes, demonstrating that our scheme is effective in reducing system latency across various scenarios. As seen from the curves in Fig.~\ref{fig:wireless}, the increase in wireless bandwidth can effectively reduce system latency, indicating the crucial role of optimized wireless bandwidth in minimizing system latency. In addition, the schemes with workload prediction outperform the schemes without workload prediction by 22\%.

The effect of various migration bandwidths on the system latency for different schemes is presented in Fig.~\ref{fig:migration}. With bandwidths of 500 Mbps, 600 Mbps, 700 Mbps, 800 Mbps, and 900 Mbps, respectively, our proposed scheme reduces up to 49\% of system latency compared to the baseline schemes, which proves that our scheme can effectively reduce the system latency across different scenarios. As can be observed from the curves in the figure, increasing migration bandwidth reduces system latency to some extent, but the impact is relatively small since a good strategy does not blindly opt to pre-migrate a large number of avatar tasks. Still, the scheme with workload prediction outperforms the scheme without workload prediction.

In the context of AeVeM, we show the impact of various RSU GPU computing resources and cloud server GPU computing resources on system latency for various schemes in Fig.~\ref{fig:edge} and Fig.~\ref{fig:cloud}. With the RSU GPU computing resources ranging from 10 GHz to 30 GHz and the cloud servers GPU computing resources ranging from 50 GHz to 100 GHz, respectively, our proposed scheme achieves the lowest system latency compared to the baseline scheme, proving that our scheme can effectively reduce the system latency across different scenarios. From the curves in the figure, we can observe that the impact of RSU GPU computing resources on system latency is greater than that of cloud servers' GPU computing resources. Moreover, the scheme with workload prediction still performs better than the scheme without workload prediction by 25\% and 27\%, respectively.


\section{Conclusions}\label{conclude}
In this paper, we have studied the avatar pre-migration problem in AeVeM. To address the challenges posed by the high mobility of intelligent vehicles, the dynamic workload of RSU, and the inhomogeneity of RSU deployment, we have formulated the problem as a POMDP and proposed a Hybrid-MAPPO algorithm capable of performing both discrete actions and continuous actions simultaneously. Furthermore, we have introduced a Coverage-aware LSTM Trajectory Prediction scheme for predicting the future load of RSUs and added it to the observations of the Hybrid-MAPPO algorithm. Numerous experimental results have demonstrated that our proposed scheme significantly outperforms the baseline schemes and can effectively reduce system latency.  There still exist some challenges to be addressed, such as security and privacy issues of avatar data migration \cite{wu2022delay}, and energy-efficient decisions for migration \cite{yang2021local}. In our future work, we consider using blockchain or other advanced privacy-preserving artificial intelligence algorithms to achieve data security and privacy protection during avatar migration in vehicular metaverses.

\bibliographystyle{IEEEtran}
\bibliography{main}

\newpage

\vfill

\end{document}